\documentclass[10pt,twocolumn,letterpaper]{article}

\usepackage[pagenumbers]{cvpr} 

\usepackage{graphicx}
\usepackage{amsmath}
\usepackage{amssymb}
\usepackage{booktabs}
\usepackage{multirow}
\usepackage{pifont}
\usepackage{amsmath}
\usepackage{overpic}
\usepackage{caption,subcaption}
\usepackage[dvipsnames,table,xcdraw]{xcolor}
\usepackage[pagebackref,breaklinks,colorlinks,linkcolor=BrickRed,citecolor=MidnightBlue,bookmarks=false]{hyperref}

\usepackage[capitalize]{cleveref}
\crefname{section}{Sec.}{Secs.}
\Crefname{section}{Section}{Sections}
\Crefname{table}{Table}{Tables}
\crefname{table}{Tab.}{Tabs.}

\def\cvprPaperID{786} 
\def\confName{CVPR}
\def\confYear{2023}

\def\ie{\emph{i.e.,~}}
\def\eg{\emph{e.g.,~}}

\newcommand{\figref}[1]{Fig.~\ref{#1}}%
\newcommand{\tabref}[1]{Tab.~\ref{#1}}%
\newcommand{\secref}[1]{Sec.~\ref{#1}}
\renewcommand{\eqref}[1]{Eqn.~(\ref{#1})}

\newcommand{\smallsec}[1]{\noindent\textbf{#1}}
\newcommand{\sota}[1]{\textbf{#1}}
\newcommand{\subsota}[1]{\underline{#1}}
\newcommand{\tablestyle}[2]{\setlength{\tabcolsep}{#1}\renewcommand{\arraystretch}{#2}\centering\footnotesize}
\definecolor{baselinecolor}{gray}{.9}
\newcommand{\baseline}[1]{\cellcolor{baselinecolor}{#1}}
\newcommand*\samethanks[1][\value{footnote}]{\footnotemark[#1]}

\begin{document}

\title{AMT: All-Pairs Multi-Field Transforms for Efficient Frame Interpolation}

\author{
  Zhen Li\thanks{Equal contribution} \quad
  Zuo-Liang Zhu\samethanks \quad
  Ling-Hao Han \quad
  Qibin Hou \quad
  Chun-Le Guo\thanks{C.L. Guo is the corresponding author.} \quad
  Ming-Ming Cheng \\
  VCIP, CS, Nankai University \\
  {\tt\small
    \{zhenli1031, nkuzhuzl\}@gmail.com,
    lhhan@mail.nankai.edu.cn
  } \\
  {\tt\small
  \{houqb, guochunle, cmm\}@nankai.edu.cn
  }
}
\maketitle

\begin{abstract}
We present \textbf{A}ll-Pairs \textbf{M}ulti-Field \textbf{T}ransforms (\textbf{AMT}),
a new network architecture for video frame interpolation.
It is based on two essential designs. 
First, we build bidirectional correlation volumes for all pairs of pixels,
and use the predicted bilateral flows to retrieve  correlations 
for updating both flows and the interpolated content feature.
Second, we derive multiple groups of fine-grained flow fields from 
one pair of updated coarse flows for performing backward warping
on the input frames separately.
Combining these two designs enables us to generate promising task-oriented flows
and reduce the difficulties in modeling large motions and handling occluded areas 
during frame interpolation.
These qualities promote our model to achieve state-of-the-art performance on various benchmarks 
with high efficiency. 
Moreover, our convolution-based model competes favorably compared to Transformer-based models in terms of accuracy and efficiency.
Our code is available at \url{https://github.com/MCG-NKU/AMT}.
\end{abstract}
\vspace{-3mm}

\section{Introduction}
\label{sec:intro}

Video frame interpolation (VFI) is a long-standing video processing technology, 
aiming to increase the temporal resolution of the input video by synthesizing 
intermediate frames from the reference ones.
It has been applied to various downstream tasks, 
including slow-motion generation~\cite{jiang2018super, xiang2020zooming}, 
novel view synthesis~\cite{zhou2016view, flynn2016deepstereo, li2021neural}, 
video compression~\cite{wu2018video}, text-to-video generation~\cite{singer2022make}, etc.

\begin{figure}
    \begin{overpic}[width=\linewidth]{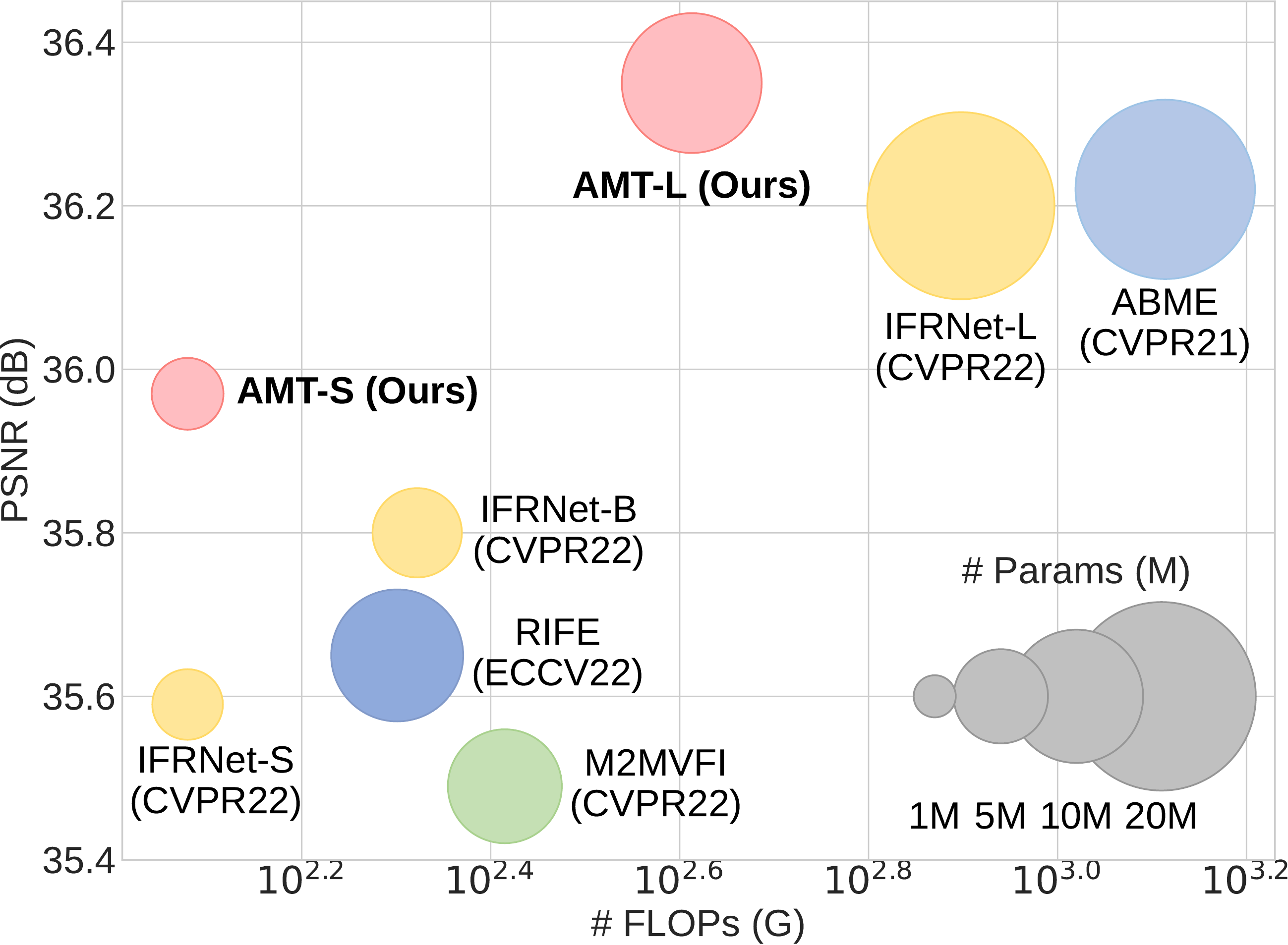}
        \put(50, 31.5){\small\linethickness{0.5mm}~\cite{Kong_2022_CVPR}}
        \put(23.5, 12){\small\linethickness{0.5mm}~\cite{Kong_2022_CVPR}}
        \put(70.8, 40){\small\linethickness{0.5mm}~\cite{Kong_2022_CVPR}}
        \put(50, 22){\small\linethickness{0.5mm}~\cite{huang2022rife}}
        \put(57, 12){\small\linethickness{0.5mm}~\cite{hu2022m2m}}
        \put(86.5, 42){\small\linethickness{0.5mm}~\cite{park2021ABME}}
    \end{overpic}
    \caption{Performance vs. number of parameters and FLOPs. 
    The PSNR values are obtained from the Vimeo90K dataset~\cite{xue2019video}.
    We use a 720p frame pair to calculate FLOPs. 
    Our AMT outperforms the state-of-the-art methods and is with higher efficiency.}
    \label{fig:comp_efficiency}
    \vspace{-5mm}
\end{figure}
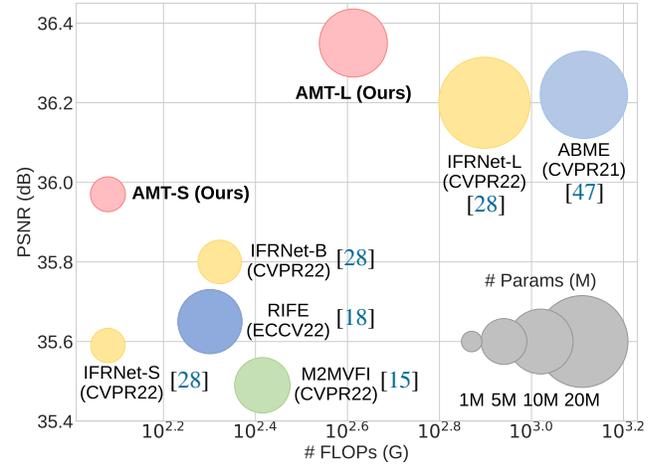
Recently, flow-based VFI methods~\cite{jiang2018super, zhang2020flexible, siyao2021anime, 
Kong_2022_CVPR, huang2022rife, lu2022vfiformer} 
have been predominant in referenced research due to their effectiveness.
A common flow-based technique estimates bilateral/bidirectional flows from the given frames 
and then propagates pixels/features to the target time step via backward~\cite{MEMC-Net, huang2022rife, Kong_2022_CVPR} 
or forward~\cite{niklaus2018context, Niklaus_CVPR_2020, hu2022m2m} warping.
Thus, the quality of a synthesized frame relies heavily on flow estimation results. 
In fact, it is cumbersome to approximate intermediate flows through pretrained optical flow models, 
and these flows are unqualified for VFI usage~\cite{hu2022m2m, huang2022rife}.

\begin{figure}
    \begin{overpic}[width=\linewidth]{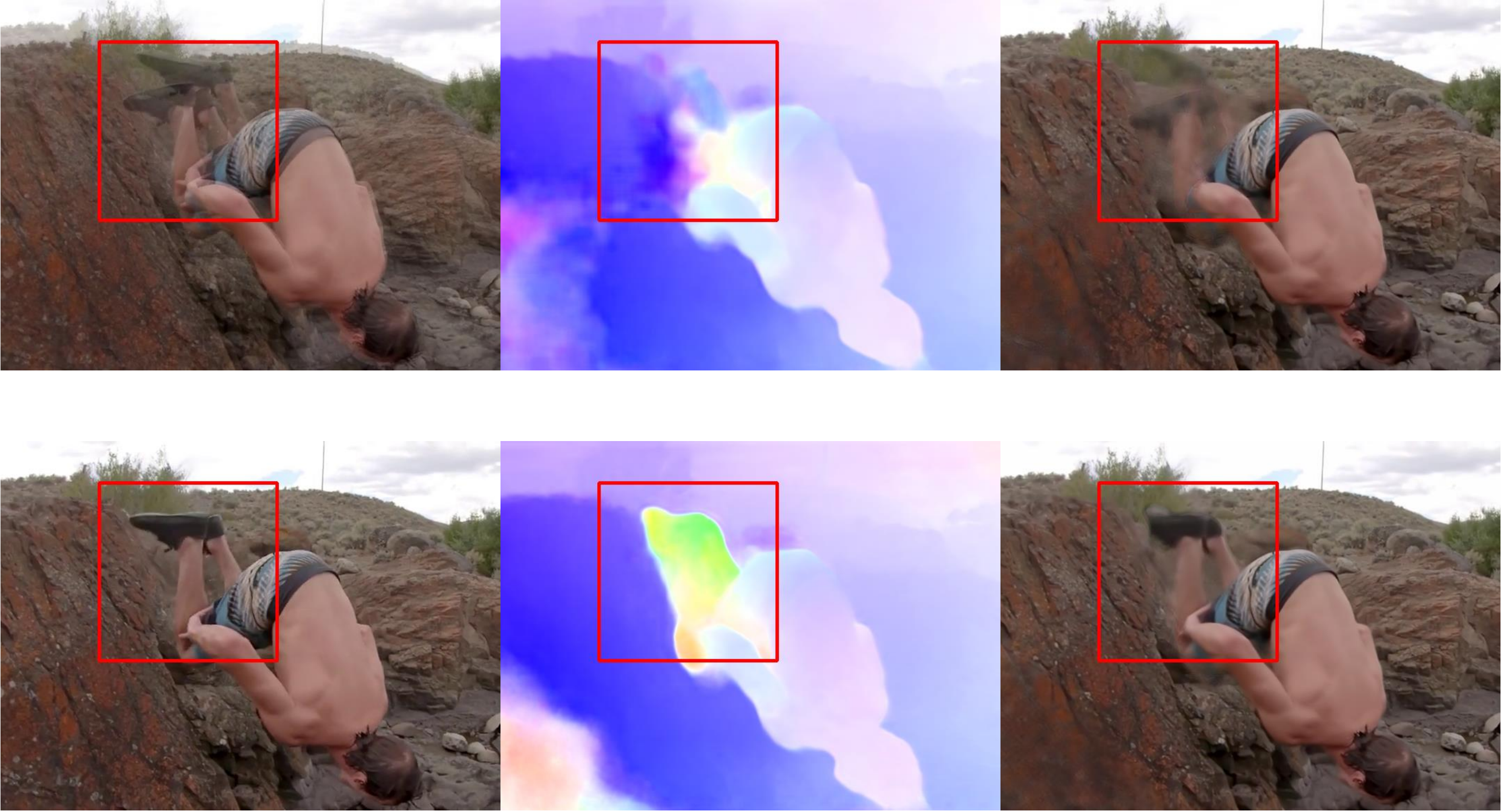}
        \put(9.8, 26.5){\small\linethickness{0.5mm}Overlaid}
        \put(36.3, 26.5){\small\linethickness{0.5mm}IFRNet~\cite{Kong_2022_CVPR} Flow}
        \put(68.4, 26.5){\small\linethickness{0.5mm}IFRNet~\cite{Kong_2022_CVPR} Result}
        \put(5.5, -2.8){\small\linethickness{0.5mm}Ground Truth}
        \put(42.0, -2.8){\small\linethickness{0.5mm}Our Flow}
        \put(75, -2.8){\small\linethickness{0.5mm}Our Result}
    \end{overpic}
    \caption{Qualitative comparisons of estimated flows and the interpolated frames. 
    Our AMT guarantees the general consistency of intermediate flows and 
    synthesizes fast-moving objects with occluded regions precisely, while
    the previous state-of-the-art IFRNet~\cite{Kong_2022_CVPR} fails to achieve them.}
    \vspace{-5mm}
    \label{fig:teaser2}
\end{figure}
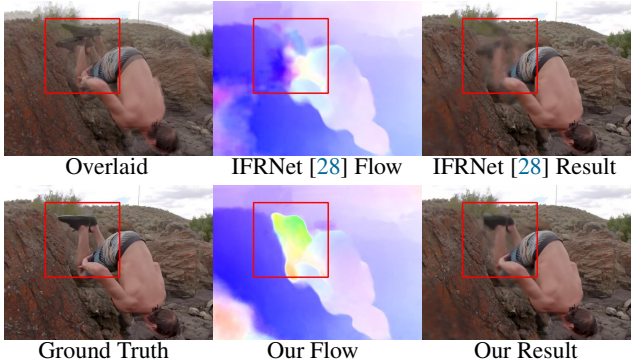

A feasible way to alleviate this issue is 
to estimate \textit{task-oriented flows} in an end-to-end training manner~\cite{liu2017voxelflow, xue2019video, jiang2018super, Kong_2022_CVPR}.
However, some major challenges, such as large motions and occlusions,
are still pending to be resolved.
These challenges mainly arise from the defective estimation of optical flows. 
Thus, a straightforward question should be: 
Why do previous methods have difficulties in predicting promising task-oriented flows when facing these challenges?
Inspired by the recent studies~\cite{xue2019video,Kong_2022_CVPR} that demonstrate \textit{task-oriented flow is generally consistent with ground truth optical flow but diverse in local details},
we attempt to answer the above question from two perspectives:

\textbf{(i)} The flow fields predicted by existing VFI methods are \textit{not consistent enough} with the true displacements, especially when encountering large motions (see \figref{fig:teaser2}).
Existing methods mostly adopt the UNet-like architecture~\cite{ronneberger2015u} with plain convolutions to build VFI models.
However, this type of architecture is vulnerable to accumulating errors at early stages when modeling large motions~\cite{teed2020raft, xu2022gmflow, zhao2022global, hou2022vfips}.
As a result, the predicted flow fields are not accurate.

\textbf{(ii)} Existing methods predict one pair of flow fields, restricting the solution set in a tight space.
This makes them struggle to handle occlusions and details around the motion boundaries,
which consequently deteriorates the final results (see \figref{fig:teaser2} and \figref{fig:occu}).
 
In this paper, we present a new network architecture, dubbed \textbf{A}ll-pairs \textbf{M}ulti-field \textbf{T}ransforms (\textbf{AMT}),
for video frame interpolation. 
AMT explores two new designs to improve the fidelity and diversity of predicted flows regarding the above two main shortcomings of previous works.

Our first design is based on all-pairs correlation in RAFT~\cite{teed2020raft},
which adequately models the dense correspondence between frames, especially for large motions.
We propose to build \textit{bidirectional correlation volumes} 
instead of a unidirectional one
and introduce a \textit{scaled} lookup strategy to 
solve the coordinate mismatch issue caused by the invisible frame.
Besides, the retrieved correlations assist our model in \textit{jointly} updating bilateral flows and the interpolated content feature in a \textit{cross-scale} manner. 
Thus, the network guarantees the fidelity of flows across scales, 
laying the foundation for the following refinement.

Second, considering that predicting one pair of flow fields is hard to cope with the occlusions, we propose to derive multiple groups of fine-grained flow fields from 
one pair of updated coarse bilateral flows.
The input frames can be separately backward warped to the target time step by these flows.
Such diverse flow fields
provide adequate potential solutions for each pixel to be interpolated, 
particularly alleviating the ambiguity issue in the occluded areas.

We examine the proposed AMT on several public benchmarks with different model scales, 
showing strong performance and high efficiency in contrast to the state-of-the-art (SOTA) methods (see \figref{fig:comp_efficiency}).
Our small model outperforms IFRNet-B, a SOTA lightweight model, 
by +0.17dB PSNR on Vimeo90K~\cite{xue2019video}
with only 60\% of its FLOPs and parameters.
For the large-scale setting, our AMT exceeds the previous SOTA (\ie IFRNet-L) 
by +0.15 dB PSNR on Vimeo90K~\cite{xue2019video}
with 75\% of its FLOPs and 65\% of its parameters.
Besides, 
we provide a huge model for comparison with the SOTA transformer-based 
method VFIFormer~\cite{lu2022vfiformer}.
Our convolution-based AMT shows a comparable performance
but only needs nearly 23$\times$ less computational cost 
compared to VFIFormer~\cite{lu2022vfiformer}.
Considering its effectiveness,
we hope our AMT could bring a new perspective
for the architecture design in efficient frame interpolation.
\vspace{-1mm}
\section{Related Work}
\smallsec{Video Frame Interpolation:}
The development of deep learning has spawned a large amount of VFI methods.
These methods can be roughly divided into three categories: 
kernel-based~\cite{cheng2020video,EDSC,lee2020adacof,Niklaus_CVPR_2017,Niklaus_ICCV_2017,peleg2019net}, 
hallucination-based~\cite{choi2020cain,kim2020fisr, kalluri2020flavr, shi2022video},
and flow-based ones~\cite{niklaus2018context, qvi_nips19, xue2019video, liu2017voxelflow, Kong_2022_CVPR, Niklaus_CVPR_2020, DAIN, MEMC-Net,huang2022rife}.

Kernel-based methods attempt to capture motion 
with dynamic kernel weights~\cite{Niklaus_CVPR_2017, Niklaus_ICCV_2017, peleg2019net} or/and 
offsets~\cite{cheng2020video, EDSC, lee2020adacof, ding2021cdfi}.
With the help of off-the-shelf architectures~\cite{shi2016real, dai2017deformable, hu2018squeeze, tran2015learning},
hallucination-based methods directly generate the interpolated frame from features of input pairs.
Thanks to the robustness of optical flow, 
flow-based methods have become mainstream in VFI.
Previous methods resort to a pretrained flow model~\cite{niklaus2018context, qvi_nips19} or 
a jointly trained estimation module~\cite{xue2019video, liu2017voxelflow, Kong_2022_CVPR}
to obtain the flow estimation.
For generating task-oriented flows, 
some methods~\cite{Kong_2022_CVPR, huang2022rife} propose intermediate supervisions 
to distill motion knowledge from the pseudo ground truth.
Subsequently, 
backward warping~\cite{MEMC-Net, huang2022rife, Kong_2022_CVPR} and 
forward warping~\cite{niklaus2018context, Niklaus_CVPR_2020, niklaus2021revisiting} are standard schemes 
in the usage of estimated flows. 
A UNet-like architecture is a common choice~\cite{niklaus2018context, Niklaus_CVPR_2020, DAIN, MEMC-Net} 
to obtain the final synthesized frame, and the transformer~\cite{vaswani2017attention, liu2021swin}, 
as a prevailing architecture, is introduced~\cite{shi2022video, lu2022vfiformer, Zhang2023ExtractingMA} for a better synthesis. 
Recent works~\cite{Kong_2022_CVPR, reda2022film} discard an independent synthetic network 
in consideration of efficiency.
However, these methods 
suffer from 
the inability in modeling large motions and in dealing with occlusions.

\begin{figure*}[t]
    \centering
    \begin{overpic}[width=\textwidth]{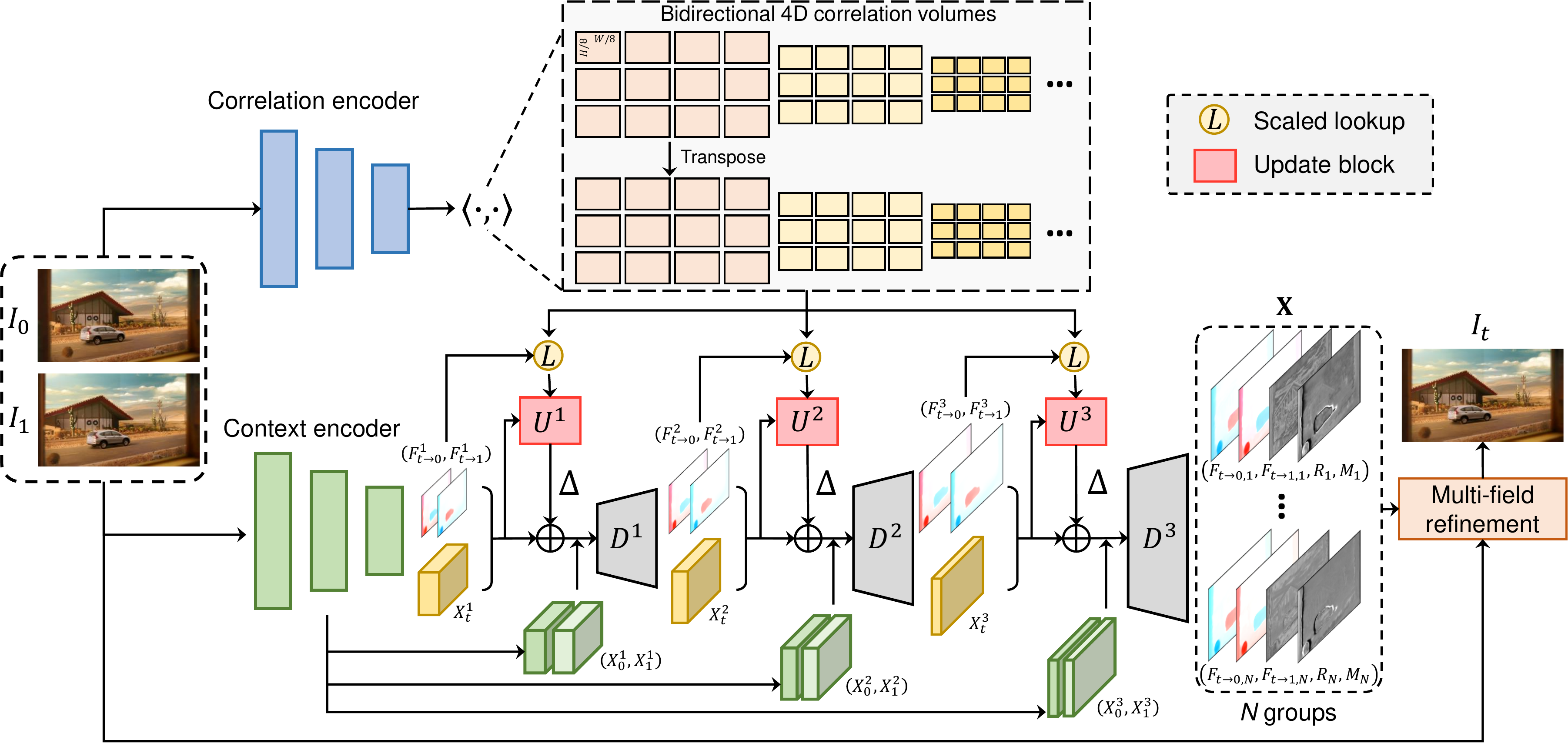}
    \vspace{-2mm}
    \end{overpic}
    \caption{Architecture overview of the proposed AMT. 
    Firstly, the input frames are sent to the correlation encoder to extract features, which are used to construct bidirectional correlation volumes.
    Then, the context encoder extracts pyramid features of visible frames and generates initial bilateral flows and interpolated intermediate feature.
    Next, we use bilateral flows to retrieve bidirectional correlations for jointly updating flow fields and the intermediate feature at each level.
    Finally, we generate multiple groups of flow fields, occlusion masks, and residuals based on the coarse estimate for interpolating the intermediate frame.
    }
    \label{fig:pipeline}
    \vspace{-14pt}
\end{figure*}

\smallsec{Task-Oriented Flow:}
Initially, flow-based video processing methods~\cite{DAIN, jiang2018super, qvi_nips19, BMBC} estimate flows and 
process images individually.
However, this two-step pipeline ignores the gap between true optical flow 
with task-specific objectives, which could be suboptimal for a specific task.
ToFlow~\cite{xue2019video} proposes the concept of task-oriented flow, 
facilitating the development of video processing 
methods~\cite{liCvpr22vInpainting, chan2022basicvsrpp, lin2022flow, yu2020joint, hu2022dmvfn} 
significantly.
Typically, the VFI-oriented flow is generally consistent with the true flow while 
diverses in detail (\eg occluded regions). 
Super Slomo~\cite{jiang2018super} introduces a mask to handle the occlusion explicitly and 
provides a standard formulation for synthesizing intermediate frames, 
which utilizes by following methods~\cite{chi2022error, Kong_2022_CVPR, 
sim2021xvfi, huang2022rife} up to now. 
IFRNet~\cite{Kong_2022_CVPR} and RIFE~\cite{huang2022rife} propose task-oriented flow distillation losses 
to provide a prior of intermediate flow in training.
Different from them,
we consider the estimation of task-oriented flows
from \textit{the perspective of architecture design}. 
We introduce all-pairs correlation to strengthen the ability 
in motion modeling, 
which guarantees the consistency of flows on the coarse scale.
At the finest scale,
we employ multi-field refinement 
to ensure the diversity for the flow regions that need to be task-specific.

\smallsec{Cost Volume:}
Cost volume is introduced as a representation of matching costs 
in numerous vision tasks~\cite{hosni2012fast, kendall2017end, jeon2018parn}.
In the deep learning era, 
the concept of cost volume is also proved to be effective 
in optical flow estimation~\cite{sun2018pwc, Hur:2019:IRR, 
teed2020raft, Zhang2021SepFlow, huang2022flowformer}.
Among these works, the most influential ones are PWC-Net~\cite{sun2018pwc} 
and RAFT~\cite{teed2020raft}.
In VFI, the existing methods~\cite{BMBC, park2021ABME, xin2022enhanced, jia2022neighbor} 
attempt to introduce the cost volume following the scheme of PWC-Net.
However, those methods not only search the cost volume in a local region
but also depend on inaccurate features warped from reference ones,
resulting in a limited performance gain from the cost volume.
Instead, the proposed AMT is based on RAFT, which enlarges the search space 
by iteratively updating the flow field with all-pairs correlation, 
and only constructs cost volumes between the visible frames.
Besides, we involve many \textit{novel} and \textit{task-specific} designs beyond RAFT.
The details are described in \secref{sec:method} and our supplement. 
\section{Method}
\label{sec:method}
Given a pair of input frames $(I_{0}, I_{1})$, 
we aim to synthesize an intermediate frame $I_{t}$
 at a target time step $t$, where $0<t<1 $.
Our AMT is a one-stage flow-based method, in which bilateral flows and the interpolated intermediate feature
are updated and upsampled jointly.
As shown in \figref{fig:pipeline}, it is composed of three main components:
1) an encoder for extracting features 
and initial bilateral flows simultaneously,
2) multi-scale bidirectional correlation volumes 
for jointly updating bilateral flows and intermediate features at coarse scales,
and 3) a multi-field refinement operator
for interpolating the target frame 
with multiple flow groups at the finest scale.
Benefiting from such designs,
the estimated motion vectors at coarser scales 
are generally consistent with the ground truth displacements.
Meanwhile, they are diverse in fine-grained details at the finest scale,
which meets the requirement of \textit{task-oriented} flow.
These designs also enable our AMT to capture large motions and 
successfully handle occlusion regions with high efficiency.

\subsection{Initial Flow and Feature Extraction}

We employ two separate feature extractors.
They are applied to the input pair $(I_{0}, I_{1})$, but for different purposes.
The first is the \textit{correlation encoder},
 which maps the input frames to a pair of dense features
 for constructing bidirectional correlation volumes.
We can obtain the pair of features $\mathbf{g}_{0}, \mathbf{g}_{1} \in \mathbb{R}^{H/8 \times W/8 \times D} $ 
at 1/8 the input image resolution with $D$ channels. 

The second is the \textit{context encoder}, 
which outputs the initial interpolated intermediate feature $X^{1}_{t}$
and predicts the initial bilateral flows $F_{t\rightarrow 0}^{1}$ and $F_{t\rightarrow 1}^{1}$.
Their spatial resolution is the same as the output of the correlation encoder.
Besides, the pyramid features $\{X^{l}_{0}, X^{l}_{1}  \mid l \in \{1,2,3\} \}$ for frames $I_{0}, I_{1}$ 
are extracted by context encoder for further progressive warping. 
The architectural details of them can be found in our supplement.

\subsection{All-Pairs Correlation}

\smallsec{Bidirectional Correlation Volumes:}
Similar to RAFT~\cite{teed2020raft}, we compute the dot-product similarities 
between all pairs of features vectors
for constructing a 4D correlation volume.
Given the pair of features $\mathbf{g}_{0}, \mathbf{g}_{1}$,
we can obtain the correlation volume $\mathbf{C}$ through:
\begin{equation}
    \mathbf{C}_{ijkl} = \sum_{h} \mathbf{g}_{0,ijh} \cdot \mathbf{g}_{1,klh}, \quad \mathbf{C} \in \mathbb{R}^{\frac{H}{8} \times \frac{W}{8} \times \frac{H}{8} \times \frac{W}{8}}
    \label{eq:corr}
\end{equation}
For further measuring similarities across scales, 
the last two dimensions of the correlation volume are downsampled 
by a repeated 2D average pooling layer with a kernel size of 2 and a stride of 2.
We thus obtain a 4-level correlation pyramid $\{\mathbf{C}_{1}, \mathbf{C}_{2}, \mathbf{C}_{3}, \mathbf{C}_{4}\}$.

However, the correlation pyramid in RAFT is \textit{unidirectional}.
It only reflects multi-scale correspondences from $I_{0}$ to $I_{1}$.
We thereby term it as the forward correlation pyramid.
The unidirectional correspondence is insufficient for the VFI task, 
as the motions are usually asymmetric~\cite{qvi_nips19, park2021ABME}.
Instead of recomputing the matrix multiplication,
we directly transpose the correlation volume $\mathbf{C}$
to represent the correspondence in the opposite direction.
After obtaining the transposed correlation volume $\mathbf{C}^{T}$, we perform the same pooling operation
to form the backward correlation pyramid $\{\mathbf{C}^{T}_{1}, \mathbf{C}^{T}_{2}, \mathbf{C}^{T}_{3}, \mathbf{C}^{T}_{4}\}$.
Note that the bidirectional correlation volumes only need to compute once.
The compact global representations 
assist our network in being aware of large motions in an efficient way.

\smallsec{Correlation Scaled Lookup:}
After constructing the bidirectional correlation volumes,
we intend to query correlation feature maps using estimated bilateral flows 
$F_{t\rightarrow 0}^{l}$ and $F_{t\rightarrow 1}^{l}$.
In RAFT, the lookup operation can be directly performed 
since 
its estimated flow and the correlation volume share an \textit{identical} coordinate system.
For example, the motion $F_{0\rightarrow 1}$ from frame 0 to frame 1 
and the corresponding correlation volume
are all based on the coordinate system of the frame 0.
Thus, the correlation feature maps can be correctly sampled by the matched flow field. 
However, for frame interpolation, we can only build correlation volumes from visible reference frames (\ie $I_0$, $I_1$)
but estimate the flows (\ie $F_{t\rightarrow 0}^{l}$ and $F_{t\rightarrow 1}^{l}$) of an invisible intermediate frame $I_t$. 
So there exists \textit{a mismatch between coordinate systems}, which causes unfaithful correlation lookups
and further influences the updating of the flows.
A straightforward solution to this problem is transferring bilateral flows $F_{t\rightarrow 0}^{l}$ and $F_{t\rightarrow 1}^{l}$
to bidirectional flows $F_{0\rightarrow 1}^{l}$ and $F_{1\rightarrow 0}^{l}$.

To achieve this goal,
we simply scale the estimated bilateral flows 
based on locally smooth motion assumption~\cite{jiang2018super,BMBC,park2021ABME}.
Specifically, we assume the moving objects are partially overlap within a small time interval. 
Thus, the bilateral flows and bidirectional flows at the same position are 
generally consistent in direction but different in magnitude.
So the bidirectional flows $F_{0\rightarrow 1}^{l}$ and $F_{1\rightarrow 0}^{l}$ 
can be approximated by:
\begin{equation}
    F_{0\rightarrow 1}^{l} = \frac{1}{1-t} F_{t\rightarrow 1}^{l}, \quad F_{1\rightarrow 0}^{l} = \frac{1}{t} F_{t\rightarrow 0}^{l}.
\end{equation}
Subsequently, a lookup operation analogous to that in RAFT performs on bidirectional 
correlation volumes through approximated bidirectional flows.
We construct two lookup windows centered by bidirectional flows 
with a predefined radius.
The lookup operations in the windows are conducted on
all levels of the bidirectional correlation pyramids.
The retrieved bidirectional correlations are concatenated into one features map
for further jointly updating bilateral flows and the interpolated intermediate feature.

\smallsec{Updating with Retrieved Correlations:}
While RAFT updates and maintains the flow prediction at a single resolution,
we predict the bilateral flows in a coarse-to-fine manner 
following most flow-based VFI methods~\cite{niklaus2018context, Niklaus_CVPR_2020,huang2022rife, Kong_2022_CVPR}.
This is because that the features of the input pair need to be progressively warped 
based on the latest flow predictions
for generating a faithful intermediate feature.
Given the reciprocal relationship between bilateral flow fields and intermediate features in VFI task~\cite{liu2017voxelflow, huang2022rife, Kong_2022_CVPR},
we also update and upsample the intermediate feature along with the intermediate motions.

Specifically, during the update stage at each spatial level $l$, 
we employ an update block to jointly predict the residuals of the bilateral flow fields $F_{t\rightarrow 0}^{l}, F_{t\rightarrow 1}^{l}$
and the interpolated intermediate feature $X_{t}^{l}$ based on the retrieved bidirectional correlations.
In each update block, the bidirectional correlation features 
and bilateral flows are first passed through two convolutional layers.
Then, they are concatenated with the interpolated intermediate feature
and injected into two convolutional layers 
instead of a cumbersome GRU unit in RAFT.
Finally, the output features are sent to two separate heads for 
predicting bilateral flow residuals $\Delta F_{t\rightarrow 0}^{l}, \Delta F_{t\rightarrow 1}^{l}$ 
and an interpolated feature residual $\Delta X_{t}^{l}$.
Each head is formed by two convolutional layers.

Note that the spatial dimension of the retrieved correlation features 
is the same as the first two dimensions of the correlation volume (\ie $\frac{H}{8} \times \frac{W}{8}$)  
but is different from that of the intermediate features 
and motions on upper levels.
We thus need to downscale the flow fields and the intermediate feature accordingly 
before feeding them into the update block and upsample the predicted residuals for updating.
Through downscaling, the update block works at a low-resolution space, leading to promising efficiency.
The updated intermediate feature $\hat{X}_{t}^{l}$ can be formulated as: $\hat{X}_{t}^{l} = X_{t}^{l} + \Delta X_{t}^{l}$,
where $\Delta X_{t}^{l}$ is the output content residual of the update block.
The updated bilateral flows $\hat{F}_{t\rightarrow 0}^{l}, \hat{F}_{t\rightarrow 1}^{l}$ 
can be obtained following the same rule.

We employ the updated bilateral flows to warp the features $X^{l}_{0}, X^{l}_{1}$ of the input frames.
Let $\hat{X}^{l}_{0}, \hat{X}^{l}_{1}$ denote the warped features.
The warped features, the updated bilateral flows, 
and the updated intermediate feature are concatenated together
and then fed into the $l$-th decoder.
The $l$-th decoder $D^{l}$ predicts the upsampled bilateral flows ${F}_{t\rightarrow 0}^{l+1}, {F}_{t\rightarrow 1}^{l+1}$
and the intermediate feature ${X}^{l+1}_{t}$ simultaneously as follows:
\begin{equation}
    \label{eq:decoder}
   [{F}_{t\rightarrow 0}^{l+1}, {F}_{t\rightarrow 1}^{l+1}, {X}^{l+1}_{t}] = D^{l}([\hat{X}^{l}_{0}, \hat{X}^{l}_{1}, \hat{F}_{t\rightarrow 0}^{l}, \hat{F}_{t\rightarrow 1}^{l}, \hat{X}^{l}_{t}]).
\end{equation}
Specially, the \eqref{eq:decoder} does not consider 
the last decoder $D^{3}$, which is responsible for generating multiple flow fields 
and occlusion masks for task-specific usage.
The architecture details of each decoder are listed in our supplement.

\subsection{Multi-Field Refinement}

{} In flow-based VFI methods, 
the common formulation for interpolating the final intermediate frame is:
\begin{equation}
    I_{t} = M \odot \mathcal{W}(I_{0}, F_{t\rightarrow 0}) + (1 - M) \odot \mathcal{W}(I_{1}, F_{t\rightarrow 1}) + R,
    \label{eq:trad_interpolate}
\end{equation}
where $\mathcal{W}$ denotes the backward warping operation, 
$\odot$ means the element-wise multiplication.
$M$ is an estimated occlusion mask which ranges from 0 to 1.
$F_{t\rightarrow 0}$ and $F_{t\rightarrow 1}$ are final predictions of bilateral flows.
$R$ is the estimated residual.
Such formulation considers temporal consistency and occlusion reasoning,
synthesizing the intermediate frame efficiently.
However, only predicting one pair of flow fields
 ignores that each location in the occlusion areas has many potential pixel candidates,
restricting the solution set for interpolation in a tight space.

Based on previously predicted coarse flows, 
which are generally consistent with the ground truth displacements, 
we derive multiple fine-grained flow fields for task-specific usage.
We also jointly estimate a residual content and an occlusion mask for each pair of optical flow.
This process can be formulated as:
\begin{equation}
    \label{eq:multiflow_generate}
    \begin{split}
        \mathbf{X} 
        = D^{3}([\hat{X}^{3}_{0}, \hat{X}^{3}_{1}, \hat{F}_{t\rightarrow 0}^{3}, \hat{F}_{t\rightarrow 1}^{3}, \hat{X}^{3}_{t}]),\\
        \mathbf{X} = \{{F}_{t\rightarrow 0, n}, {F}_{t\rightarrow 1, n}, {M}_{n}, {R}_{n} | n \in \{1,2,..., N\}\},
    \end{split}
\end{equation}
where $N$ denotes the total number of output groups. 
$({F}_{t\rightarrow 0, n}, {F}_{t\rightarrow 1, n})$, ${M}_{n}$, and ${R}_{n}$ 
are the $n$-th estimated bilateral flows, occlusion mask, and residual content, respectively.
Notably, \eqref{eq:multiflow_generate} can be easily implemented by enlarging the output channels 
of the last decoder 
according to the number of flow pairs,
which ensures efficiency.
The final intermediate frame can be obtained by:
\begin{equation}
    \label{eq:our_interpolate}
        I_{t} = \mathcal{C} ([I_{t}^{1}, ..., I_{t}^{N}]), \\ 
\end{equation}
where the $n$-th interpolated frame $I_{t}^{n}$ can be obtained by \eqref{eq:trad_interpolate} 
with corresponding output group.
We stack two convolutional layers (denoted as $\mathcal{C}$) 
for adaptively merging candidate frames and refining the final results. 
The analyses of multiple flow fields are detailed in \secref{sec:multifield}.

\subsection{Loss Functions}
There are three losses involved in our AMT.
To better predict task-oriented flows,
we employ flow distillation loss $\mathcal{L}_{flow}$ in IFRNet~\cite{Kong_2022_CVPR}, 
which concentrates more on the flow regions that are easy to be reconstructed,
but slightly penalizes the regions that are difficult to recover.
This loss is applied on updated multi-scale flow fields
except for the finest flow predictions left for fully task-specific usage.
The Charbonnier loss~\cite{charbonnier1994two} $\mathcal{L}_{char}$
 and the census loss~\cite{Meister:2018:UUL} $\mathcal{L}_{css}$ are used to supervise the content generation of the interpolated frame.
The former measures the pixel-wise errors between the ground truth intermediate frame $I_{t}^{GT}$ and the generated one $I_{t}$, 
and the latter calculates the soft Hamming distance 
between census-transformed image patches of $I_{t}^{GT}$ and $I_{t}$.

The full objective can be defined as:
\begin{equation}
    \mathcal{L} = \lambda_{char}\mathcal{L}_{char} + \lambda_{css}\mathcal{L}_{css} + \lambda_{flow}\mathcal{L}_{flow},
\end{equation}
where $\lambda_{char}$, $\lambda_{css}$, and $\lambda_{flow}$ are weights for each loss.
\begin{table*}[ht]
    \small
    \tablestyle{2pt}{1}
    \begin{tabular}{lccccccccccc}
\toprule
    \multirow{2}{*}{Method} & \multirow{2}{*}{Vimeo90K~\cite{xue2019video}} & \multirow{2}{*}{UCF101~\cite{soomro2012ucf101}} & \multicolumn{4}{c}{SNU-FILM~\cite{choi2020cain}} & \multicolumn{2}{c}{Xiph~\cite{xiph1994}} & Latency & Params & FLOPs \\
\cmidrule(lr){4-7} \cmidrule(lr){8-9}
    &  &  & Easy & Medium & Hard & Extreme & 2K & 4K & (ms/f) & (M) & (T) \\
\midrule
AdaCoF~\cite{lee2020adacof}    & 34.38/0.972 & 35.20/0.970 & 39.85/0.991 & 35.08/0.976 & 29.47/0.925 & 24.31/0.844 & 34.86/0.928 & 31.68/0.870 & 52 & 21.8 & 0.36 \\
M2M-VFI~\cite{hu2022m2m}   & 35.49/0.978 & \subsota{35.32/0.970} & 39.66/\subsota{0.991} & 35.74/\subsota{0.980} & 30.32/\subsota{0.936} & 25.07/0.860  & \sota{36.44}/\sota{0.943} & 33.92/0.899 & 40 & 7.6  & 0.26 \\
RIFE~\cite{huang2022rife}      & 35.65/0.978 & 35.28/0.969 & \sota{40.06}/0.991 & 35.75/0.979 & 30.10/0.933 & 24.84/0.853 & 36.19/0.938 & 33.76/0.894 & 29 & 9.8  & 0.20 \\
IFRNet-S~\cite{Kong_2022_CVPR}  & 35.59/0.979 & 35.28/0.969 & \subsota{39.96}/0.991 & 35.92/0.979 & 30.36/0.936 & 25.05/0.858 & 35.87/0.936 & 33.80/0.891 & 25 & 2.8  & 0.12 \\
IFRNet-B~\cite{Kong_2022_CVPR}  & \subsota{35.80/0.979} & 35.29/0.969 & 40.03/0.991 & \subsota{35.94}/0.979 & \subsota{30.41}/0.936 & \subsota{25.05/0.859} & 36.00/0.936 & \subsota{33.99/0.893} & 30 & 5.0  & 0.21 \\
AMT-S             & \sota{35.97/0.983} & \sota{35.35/0.971} & 39.95/\sota{0.994} & \sota{35.98/0.983} & \sota{30.60/0.940} & \sota{25.30/0.865} &\subsota{36.11}/\subsota{0.940} & \sota{34.29}/\sota{0.901} & 51 & 3.0 & 0.12 \\
\midrule
    ToFlow~\cite{xue2019video}     & 33.73/0.968 & 34.58/0.967 & 39.08/0.989 & 34.39/0.974 & 28.44/0.918 & 23.39/0.831 & 33.93/0.922 & 30.74/0.856 & 88 & 1.4 & 0.62  \\
    DAIN~\cite{DAIN}               & 34.71/0.976 & 34.99/0.968 & 39.73/0.990 & 35.46/0.978 & 30.17/0.934 & 25.09/0.858 & 35.95/0.940 & 33.49/0.895 & 664 & 24.0 & 5.51 \\
    CAIN~\cite{choi2020cain}       & 34.78/0.974 & 35.00/0.969 & \subsota{39.95}/0.990 & 35.66/0.978 & 29.93/0.930 & 24.80/0.851 & 35.21/0.937 & 32.56/0.901 & 71 & 42.8 & 1.29 \\
    BMBC~\cite{BMBC}                   & 35.01/0.976 & 35.15/0.969 & 39.90/0.990 & 35.31/0.977 & 29.33/0.927 & 23.92/0.843 & 32.82/0.928 & 31.19/0.880 & 2234 & 11.0 & 2.50 \\
    ABME~\cite{park2021ABME}       & \subsota{36.22/0.981} & 35.41/\subsota{0.970} & 39.59/0.990 & 35.77/0.979 & 30.58/0.937 & \sota{25.42/0.864} & \sota{36.53}/\sota{0.944} & 33.73/\subsota{0.901} & 560 & 18.1 & 1.30 \\
    IFRNet-L~\cite{Kong_2022_CVPR} & 36.20/0.981 & \subsota{35.42}/0.970 & \sota{40.10}/\subsota{0.991} & \sota{36.12}/\subsota{0.980} & 30.63/0.937 & 25.27/0.861 & 36.21/0.937 & \subsota{34.25}/0.895 & 80 & 19.7 & 0.79 \\    
    AMT-L                          & \sota{36.35/0.982} & \sota{35.42/0.970} & 39.95/\sota{0.991} & \subsota{36.09}/\sota{0.981} & \sota{30.75/0.938} & \subsota{25.41/0.864}  & \subsota{36.27}/\subsota{0.940} & \sota{34.49}/\sota{0.903} & 116 & 12.9 & 0.58 \\
\midrule
    VFIFormer~\cite{lu2022vfiformer} & \subsota{36.50/0.982} & \subsota{35.43/0.970} & \sota{40.13}/0.991 & 36.09/0.980 & 30.67/0.938 & 25.43/0.864  & OOM & OOM & 1293 & 24.1 & 47.71 \\
    EMA-VFI$^{\dag}$~\cite{Zhang2023ExtractingMA}  & 36.50/0.980 & 35.42/0.970 & 39.58/0.989 & 35.86/0.979 & \sota{30.80}/0.938 & \sota{25.59}/0.864  & \sota{36.74}/\sota{0.944} & \subsota{34.55}/\sota{0.906} & 211 & 66.0  & 0.91 \\
    AMT-G & \sota{36.53}/\sota{0.982} & \sota{35.45}/\sota{0.970} & \subsota{39.88}/\sota{0.991} & \sota{36.12}/\sota{0.981} & \subsota{30.78}/ \sota{0.939} & \subsota{25.43} /\sota{0.865}  & \subsota{36.38}/\subsota{0.941} & \sota{34.63}/\subsota{0.904} & 250 & 30.6 & 2.07 \\
\bottomrule
    \end{tabular}
    \vspace{-8pt}
    \caption{Quantitative comparison with SOTA methods.
    We divide the existing methods into three groups, according to the computational complexity. 
    For each group, the best result is shown in \sota{bold}, and the second best is \subsota{underlined}. ``OOM" denotes the out-of-memory issue when evaluating on an NVIDIA RTX 3090 GPU.
    $^{\dag}$ means we disable the test-time augmentation~\cite{huang2022rife} for a fair comparison. 
    }
    \label{tab:res}
    \vspace{-10pt}
\end{table*}

\vspace{-5pt}
\section{Experiments}

\vspace{-2pt}
\subsection{Training Details}
\vspace{-2pt}

We train AMT on Vimeo90K~\cite{xue2019video} training set for 300 epochs 
with AdamW~\cite{loshchilov2018decoupled} optimizer on 2 NVIDIA RTX 3090 GPUs. 
The total batch size is 24, and the learning rate decay follows 
the cosine attenuation schedule from $2\times 10^{-4}$ to $2\times 10^{-5}$.
We follow the augmentation pipeline
including random flipping, rotating, reversing sequence order, 
and random cropping patches with size $224\times 224$ in IFRNet~\cite{Kong_2022_CVPR}. 
The flow predictions from the pre-trained LiteFlowNet~\cite{hui18liteflownet} are served as 
the pseudo ground truth label for supervising the intermediate flows.
$\lambda_{char}$, $\lambda_{css}$, and $\lambda_{flow}$ are set as 1, 1, and 0.002, respectively.
The code implemented by MindSpore framework is also provided.
\vspace{-2pt}
\subsection{Benchmarks}
\vspace{-2pt}
We evaluate our AMT on various benchmarks containing 
diverse motion scenes for a comprehensive comparison. 
PSNR and SSIM~\cite{wang2004image} as common evaluation metrics are 
utilized for comparison. 
The statistics of benchmarks used in the main paper are presented as follows.

\noindent
\textbf{Vimeo90K}~\cite{xue2019video}: 
Vimeo90K is the most commonly used evaluation benchmark 
in recent VFI literature.
There are 3,782 triplets of $448\times256$ resolution in the test part.

\noindent
\textbf{UCF101}~\cite{soomro2012ucf101}:
UCF101 dataset contains videos with various human actions, and 
we adopt the test partition in DVF~\cite{liu2017voxelflow}, 
which consists of 379 triplets of $256\times256$ size.

\noindent
\textbf{SNU-FILM}~\cite{choi2020cain}:
SNU-FILM dataset contains 1,240 frame triplets, whose width 
ranges from 368 to 720 and height ranges from 384 to 1280. 
With respect to motion magnitude, it is partitioned into four 
exclusive parts, namely Easy, Medium, Hard, and Extreme.

\noindent
\textbf{Xiph}~\cite{xiph1994}:
Xiph dataset, consisting of eight video clips with a 4K resolution, was originally proposed by Niklaus \textit{et al}.~\cite{Niklaus_CVPR_2020}. Following their original evaluation setting, we reform this dataset to include ``2K" version, obtained by downscaling original frames, and ``4K" version, created by center-cropping 2K patches.

Except for these datasets, we provide the comparisons of multi-frame interpolation in the supplement.

\vspace{-2pt}
\subsection{Comparison with the SOTAs}
\vspace{-2pt}
We compare our AMT with the state-of-the-art (SOTA) methods, 
including 
ToFlow~\cite{xue2019video}, DAIN~\cite{DAIN}, CAIN~\cite{choi2020cain}, 
AdaCoF~\cite{lee2020adacof}, BMBC~\cite{BMBC}, 
RIFE~\cite{huang2022rife},
ABME~\cite{park2021ABME}, 
M2M-VFI~\cite{hu2022m2m}, 
IFRNet~\cite{Kong_2022_CVPR},
VFIFormer~\cite{lu2022vfiformer},
and EMA-VFI~\cite{Zhang2023ExtractingMA}.
We utilize the code provided by IFRNet~\cite{Kong_2022_CVPR} for benchmarks.
The inference latency is the average running time of 
a method on $1280\times 720$ resolution for 1000 iterations on an NVIDIA RTX 3090 GPU.
To ensure a fair comparison, we group the SOTA methods into three categories based on their theoretical computational complexity. 
We then develop three models, called AMT-S, AMT-L, and AMT-G, for each group.

\smallsec{Quantitative Comparison.} 
As shown in \tabref{tab:res},
our small model AMT-S achieves the best results among 
efficient VFI methods on almost all benchmarks, 
especially for challenging settings.
Specifically, Our AMT-S outperforms the previous state-of-the-art method in effective VFI, IFRNet-B~\cite{Kong_2022_CVPR}, by 0.17dB on Vimeo90K while using only about 60\% of its parameters and FLOPs.
This gap becomes more obvious on the Hard and Extreme partitions in SNU-FILM,
revealing the strong ability of our AMT in modeling large motions. 
For the large scale setting, our AMT-L shows highly competitive results 
in contrast to the previous SOTA method IFRNet-L~\cite{Kong_2022_CVPR},
with about 65\% parameters and 75\% FLOPs of it. 
In terms of inference speed, our method is comparable to IFRNet.
Besides, our convolution-based model competes favorably compared to the SOTA Transformer-based models (\ie VFIFormer~\cite{lu2022vfiformer} and EMA-VFI~\cite{Zhang2023ExtractingMA}) in terms of accuracy and efficiency.
Specifically, our AMT-G outperforms them in most casess, particularly when evaluated using the SSIM metric.
Notably, our model achieves about 5$\times$ faster inference speed than VFIFormer and has only half the number of parameters of EMA-VFI.
It is important to note that VFIFormer requires a two-stage training pipeline and 600 training epochs, while our model only requires 300 epochs. 
Additionally, EMA-VFI introduces a warm-up technique during training, which our method does not utilize.
We observe that the performance of our method is saturated except for the Vimeo90K dataset
after increasing the scale of the model to a huge version, which may indicate the overfitting problem.

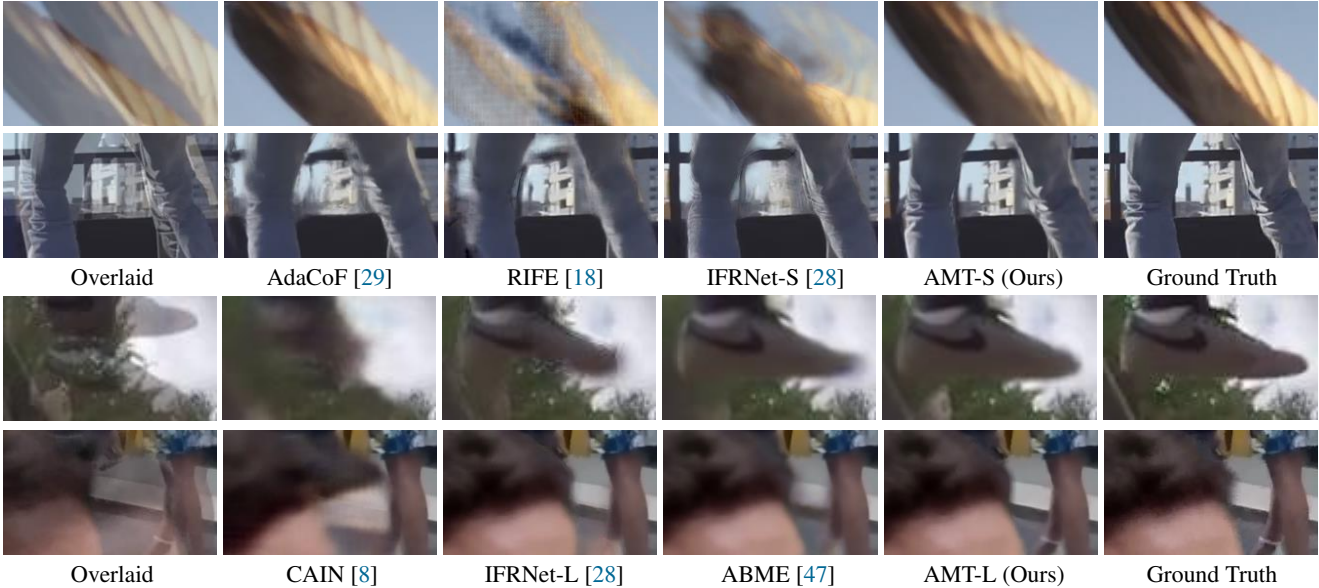
\begin{figure*}[t]
    \vspace{-5pt}
    \begin{overpic}[width=\linewidth]{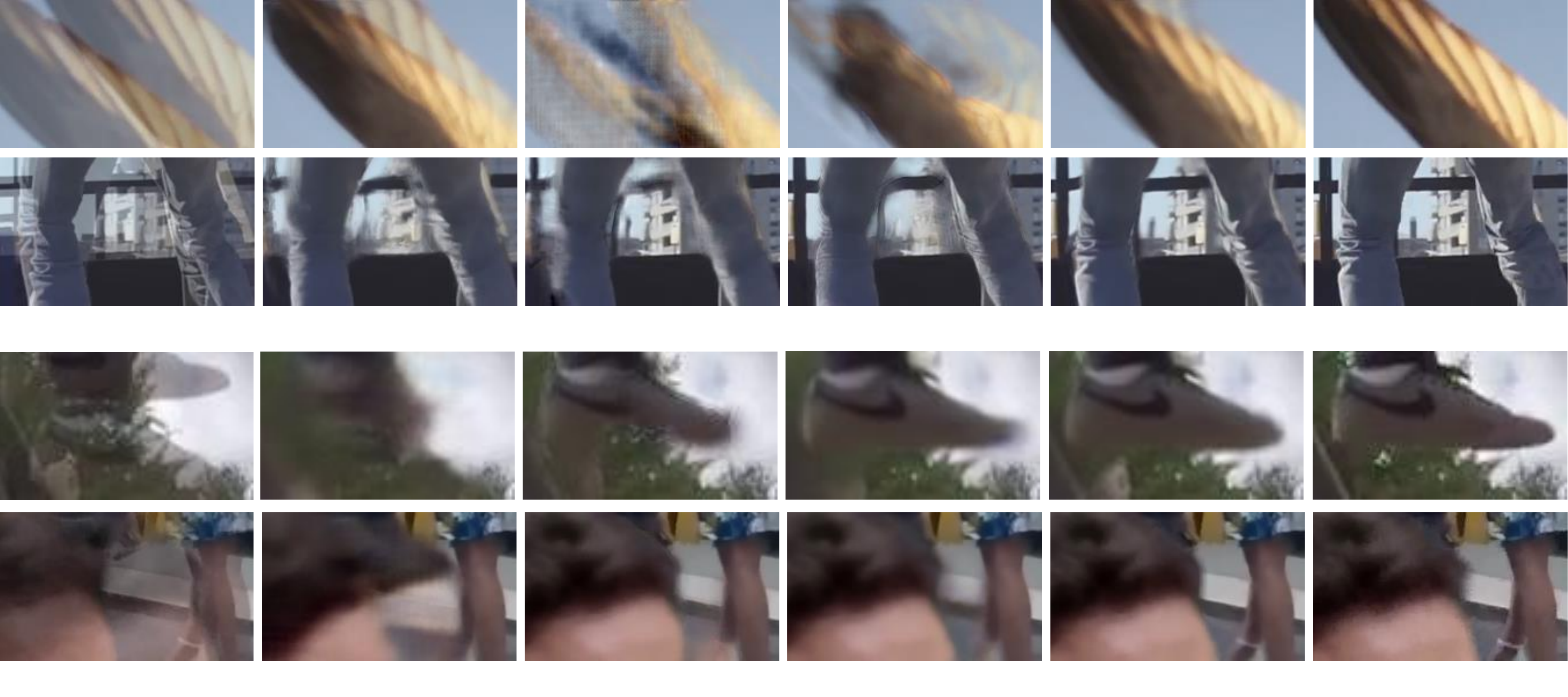}
        \put(5.1, 22.9){\small\linethickness{0.5mm}Overlaid}
        \put(20.0, 22.9){\small\linethickness{0.5mm}AdaCoF~\cite{lee2020adacof}}
        \put(38.3, 22.9){\small\linethickness{0.5mm}RIFE~\cite{huang2022rife}}
        \put(53.5, 22.9){\small\linethickness{0.5mm}IFRNet-S~\cite{Kong_2022_CVPR}}
        \put(70, 22.9){\small\linethickness{0.5mm}AMT-S (Ours)}
        \put(87.0, 22.9){\small\linethickness{0.5mm}Ground Truth}
        \put(5.1, 0.25){\small\linethickness{0.5mm}Overlaid}
        \put(21.5, 0.25){\small\linethickness{0.5mm}CAIN~\cite{choi2020cain}}
        \put(36.6, 0.25){\small\linethickness{0.5mm}IFRNet-L~\cite{Kong_2022_CVPR}}
        \put(54.6, 0.25){\small\linethickness{0.5mm}ABME~\cite{park2021ABME}}
        \put(70, 0.25){\small\linethickness{0.5mm}AMT-L (Ours)}
        \put(87.0, 0.25){\small\linethickness{0.5mm}Ground Truth}
    \end{overpic}
    \caption{Qualitative results from different VFI methods. We divide these methods 
    into two groups by computational cost.
    Our AMT-S and AMT-B synthesize precise boundaries of the objects with large motion  and can 
    reconstruct occluded regions with high fidelity. 
    }
    \vspace{-12pt}
    \label{fig:qual}
\end{figure*}

\smallsec{Qualitative Comparison.} 
In \figref{fig:qual}, we 
select the representative hallucination-based, kernel-based, and flow-based methods, 
including CAIN~\cite{choi2020cain}, AdaCoF~\cite{lee2020adacof}, 
ABME~\cite{park2021ABME}, RIFE~\cite{huang2022rife}, and
IFRNet(-B/-L)~\cite{Kong_2022_CVPR}.
We compare them with our AMT on SNU-FILM~\cite{choi2020cain} (Hard) dataset 
for visual comparison.
It can be seen that previous VFI methods fail to provide sharp edges of 
moving objects, especially when the motion is complex. 
Due to our thorough consideration of VFI-oriented flows,
our AMT synthesizes the content at motion boundaries faithfully and 
generates plausible textures with fewer artifacts. 
When the background objects are heavily occluded by the foreground unilaterally, 
our AMT can still obtain guidance from the reference frame in another direction, 
while other methods are unable to synthesize these occluded objects.
We provide more comparisons in the supplement.

\begin{table*}[t]
    \centering
    \begin{subtable}[h]{0.32\linewidth}
        \centering
        \tablestyle{4pt}{1.05}
        \begin{tabular}{lccc}
            \toprule
            Case & Vimeo & Hard & Extreme \\
            \midrule
            w/o Corr. Enc.   & 35.76 & 30.49 & 25.22 \\ 
            Unidir. CV   & 35.93 & 30.34 & 25.18 \\ 
            PWC CV   & 35.61 & 30.48 & 25.16 \\
            Full Model    & \baseline{\textbf{35.97}} & \baseline{\textbf{30.60}} & \baseline{\textbf{25.30}} \\
            \bottomrule
        \end{tabular}
        \captionsetup{width=0.9\textwidth}
        \caption{\textbf{Correlation volume (CV) design.} 
        We remove the correlation encoder (`w/o Corr. Enc.'), 
        build a unidirectional CV (`Unidir. CV'), and 
        build PWC-like~\cite{ren2018fusion} CV (`PWC CV') for ablations, respectively.}
        \label{tab:ab-cvdesign}
     \end{subtable}
     \hspace{1mm}
    \begin{subtable}[h]{0.32\linewidth}
        \centering
        \tablestyle{3.2pt}{1.05}
        \begin{tabular}{ccccc}
            \toprule
            Lookup & Init & Vimeo & Hard & Extreme \\
            \midrule
            \multicolumn{2}{c}{Initial meshgrid} & 35.92 & 30.52 & 25.23 \\ 
            RAFT & Flow & 35.93 & 30.34 & 25.18 \\ 
            Scaled  & Zero & 35.97 & 30.56 & 25.26 \\ 
            Scaled  & Flow & \baseline{\textbf{35.97}} & \baseline{\textbf{30.60}} & \baseline{\textbf{25.30}} \\ 
            
            \bottomrule
                \end{tabular}
       \captionsetup{width=0.923\textwidth}
       \caption{\textbf{Lookup strategy}. 
       We investigate the initial meshgrid, RAFT-like~\cite{teed2020raft} lookup (`RAFT'), 
       and the proposed lookup (`Scaled') variants.
       We also investigate whether we use bilateral flows to perform an initial lookup. 
       }
       \label{tab:ab-lookup}
    \end{subtable}
    \hspace{1mm}
     \begin{subtable}[h]{0.32\linewidth}
        \centering
        \tablestyle{4pt}{1.05}
            \begin{tabular}{lccc}
        \toprule
        Case & Vimeo & Hard & Extreme \\
        \midrule
        Vanilla Guide & 35.95 & 30.53 & 25.21 \\ 
        w/o Update    & 35.96 & 30.52 & 25.22 \\ 
        Full Model    & \baseline{\textbf{35.97}} & \baseline{\textbf{30.60}} & \baseline{\textbf{25.30}} \\
        \bottomrule
        \end{tabular}
        \captionsetup{width=0.95\textwidth}
        \caption{\textbf{Content update.}
        We investigate the content update by using features from visible frames 
        as guidance (`Vanilla Guide') and discarding the content update (`w/o Update'), 
        respectively.}
        \label{tab:ab-contupdate}
     \end{subtable} \\
     \centering
     \begin{subtable}[h]{0.32\linewidth}
        \centering
        \tablestyle{3.5pt}{1.05}
            \begin{tabular}{cccccc}
                \toprule
                1st & ~2nd & 3rd & Vimeo & Hard & Extreme \\
                \midrule
                          &           &           & 35.60 & 30.39 & 25.06 \\
                \ding{52} &           &           & 35.84 & 30.55 & 25.19 \\
                \ding{52} & \ding{52} &           & 35.92 & 30.58 & 25.28 \\
                \ding{52} & \ding{52} & \ding{52} & \baseline{\textbf{35.97}} & \baseline{\textbf{30.60}} & \baseline{\textbf{25.30}} \\
                \cline{1-3}
                \multicolumn{3}{c}{single-scale}  & 35.95 & 30.50 & 25.22 \\ 
                \bottomrule
            \end{tabular}
            \captionsetup{width=0.98\textwidth}
            \caption{\textbf{Cross-scale update}. We investigate the impact of the
        update at different levels.}
        \label{tab:ab-update}
     \end{subtable}
     \hspace{1mm}
    \begin{subtable}[h]{0.32\linewidth}
        \centering
        \tablestyle{2pt}{1.05}
            \begin{tabular}{cccccc}
                \toprule
                No. & Vimeo & Hard & Extreme & FLOPs (G) \\
                \midrule
                1   & 35.84 & 30.52 & 25.25 & 116 \\
                3  & \baseline{35.97} & \baseline{30.60} & \baseline{25.30} & \baseline{121} \\
                5 & 36.00 & \textbf{30.63} & \textbf{25.33} & 127 \\
                7 & \textbf{36.01} & 30.57 & 25.25 & 135 \\
                \bottomrule
                \end{tabular}
        \captionsetup{width=0.85\textwidth}
        \caption{\textbf{Number of fields}. We investigate different numbers 
        of flow pairs.}
        \label{tab:ab-flo}
     \end{subtable}
     \hspace{1mm}
    \begin{subtable}[h]{0.32\linewidth}
        \centering
        \tablestyle{4pt}{1.05}
            \begin{tabular}{lccc}
                \toprule
                Case & Vimeo & Hard & Extreme \\
                \midrule
                w/o Residual  & 35.87 & 30.57 & 25.27 \\ 
                w/o Refine     & 35.89 & 30.51 & 25.19 \\ 
                Full Model   & \baseline{\textbf{35.97}} & \baseline{\textbf{30.60}} & \baseline{\textbf{25.30}} \\
                \bottomrule
            \end{tabular}
            \captionsetup{width=0.9\textwidth}
            \caption{\textbf{Multi-field combination.} 
        We investigate the residual component in \eqref{eq:trad_interpolate} and the refinement step in \eqref{eq:our_interpolate}.}
        \label{tab:ab-flowres}
     \end{subtable}
     \vspace{-3.5mm}
     \caption{Ablation experiments of AMT on Vimeo90K~\cite{xue2019video} and SNU-FILM~\cite{choi2020cain} (Hard, Extreme) dataset. 
                We report the PSNR values of these variants, and the best result is shown in \textbf{bold}.
                The default setting is marked in \colorbox{baselinecolor}{gray}.
                }
    \vspace{-15pt}
\end{table*}

\vspace{-2pt}
\subsection{Ablation Study}
\vspace{-2pt}

We conduct ablations to verify the effectiveness of two key components 
(\ie all-pairs correlation and multi-field refinement) in our AMT.
All ablated versions are based on the AMT-S 
and evaluated on Vimeo90K~\cite{xue2019video} 
and the Hard and Extreme partitions of SNU-FILM~\cite{choi2020cain}.

\vspace{-10pt}
\subsubsection{All-Pairs Correlation}

\smallsec{Volume Designs.}
As illustrated in \tabref{tab:ab-cvdesign}, our bidirectional correlation volumes 
thoroughly consider the correspondences between input frames
for the VFI task, leading to a better performance than the unidirectional one.
Besides,
using an exclusive encoder (\ie correlation encoder) for building the correlation volumes is necessary.
We can observe that the performance heavily drops 
when we utilize features from the context encoder to construct the correlation volumes.
We also try to build the correlation volumes following PWC-Net~\cite{sun2018pwc}. 
This variant performs worse than any other one, 
for its partial correlation volume limits the ability in modeling motion information sufficiently.

\smallsec{Lookup Strategy.}
As shown in \tabref{tab:ab-lookup}, 
we can observe an obvious performance drop 
while utilizing the vanilla lookup strategy in RAFT~\cite{teed2020raft}.
For large motions, 
its performance is even worse than the one that directly uses the initial meshgrid, 
which indicates this strategy provides unfaithful correlation information 
for flow updates.
After we project the flows by scaling, 
the correlation volumes and flows share an identical coordinate system, and 
the network takes advantage of the correct lookup process.
Besides, 
the initial flow pair from the context encoder gives a good initial point for further lookup, 
which brings a performance gain.

\smallsec{Content Update.}
In our AMT, each update block receives 
the intermediate content features as the context guidance 
and updates it along with bilateral flows.
If we replace the context guidance with features from visible frames,
the ambiguous information will be introduced, 
leading to a performance drop, as shown in \tabref{tab:ab-contupdate}.
Besides, we only keep one head in each update block for 
only updating the flow fields without updating the intermediate feature,
resulting in the decrease of PSNR values on large motions.
It demonstrates that all-pairs correlation is not only helpful for updating flows but also for updating content.

\smallsec{Update Strategy.}
As shown in \tabref{tab:ab-update},  
all updates across levels are effective in our cross-scale update strategy.
It is worth noticing that if we discard all updates, 
which is equivalent to a model without all-pairs correlation,
the PSNR value will decrease dramatically.
This demonstrates \textit{the effectiveness of all-pairs correlation} in our AMT.
Besides, 
only updating on the 1-st scale with $3\times$ iterations degrades the performance.
The fact indicates that the cross-scale update strategy can take full advantage 
of progressively refined content features, leading to better motion modeling.

\vspace{-10pt}
\subsubsection{Multi-field Refinement}
\label{sec:multifield}

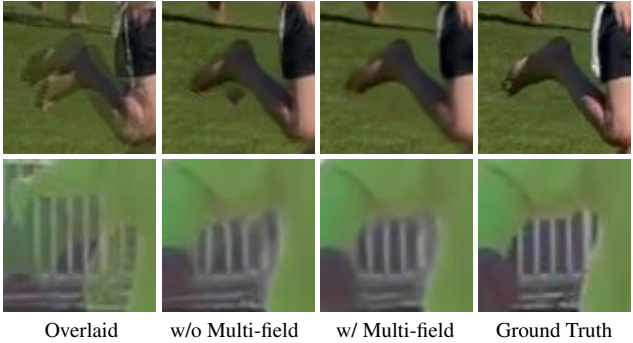
\begin{figure}[t]
    \begin{overpic}[width=\linewidth]{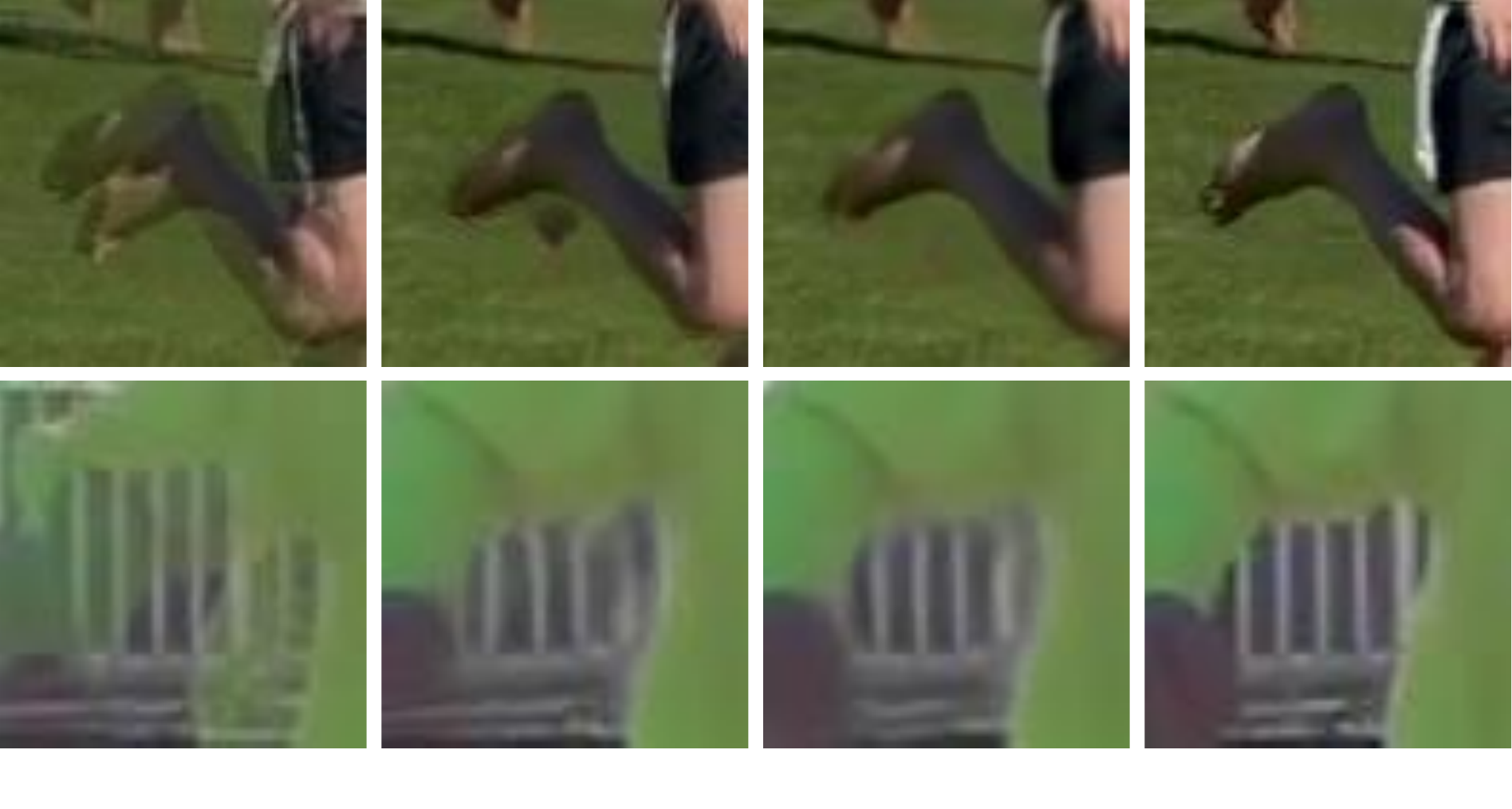}
        \put(6.5,0){\footnotesize\linethickness{0.5mm}Overlaid}
        \put(26.5,0){\footnotesize\linethickness{0.5mm}w/o Multi-field}
        \put(53.0,0){\footnotesize\linethickness{0.5mm}w/ Multi-field}
        \put(78.5,0){\footnotesize\linethickness{0.5mm}Ground Truth}
    \end{overpic}
    \caption{Effect of multi-field refinement. 
    Multi-field refinement helps the network recover occluded regions better.
    }
    \label{fig:occu}
    \vspace{-15pt}
\end{figure}

\vspace{-1mm}
\smallsec{Number of Flow Fields.} 
\tabref{tab:ab-flo} illustrates the performance gain with respect to 
the number of flow fields.
We observe that just using three pairs of flows bring a notable performance gain, 
which reveals that ensuring the diversity of flow fields is significant for VFI-oriented usage.
The PSNR values rises in pace with the increase of field number 
until 7 pairs, which indicates saturation. 
We employ 3 pairs in our small model for efficiency (\ie AMT-S) and use 5 pairs in the larger models for better performance.
In \figref{fig:occu}, we investigate the effect of multi-field refinement on occlusion handling.
The results indicate that after employing multi-field refinement, our AMT can synthesize 
the background occluded by the foreground with more consistent textures.

\smallsec{Multi-Field Combination.} 
We investigate a variant that removes the residual component for each candidate frame in \eqref{eq:trad_interpolate}
 but estimates the residual part in the final interpolation result.
As shown in \tabref{tab:ab-flowres}, the results of this variant underperform the original setting,
which indicates we need to compensate details for each frame candidate separately.
Besides, if we replace the convolution operators in \eqref{eq:our_interpolate} 
with an average operation, the performance will be degraded (see \tabref{tab:ab-flowres}).
This indicates that it is important for our AMT to perform an adaptive fusion and refinement.

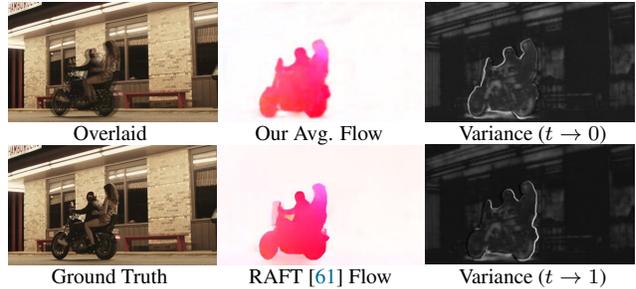
\begin{figure}[t]
    \begin{overpic}[width=\linewidth]{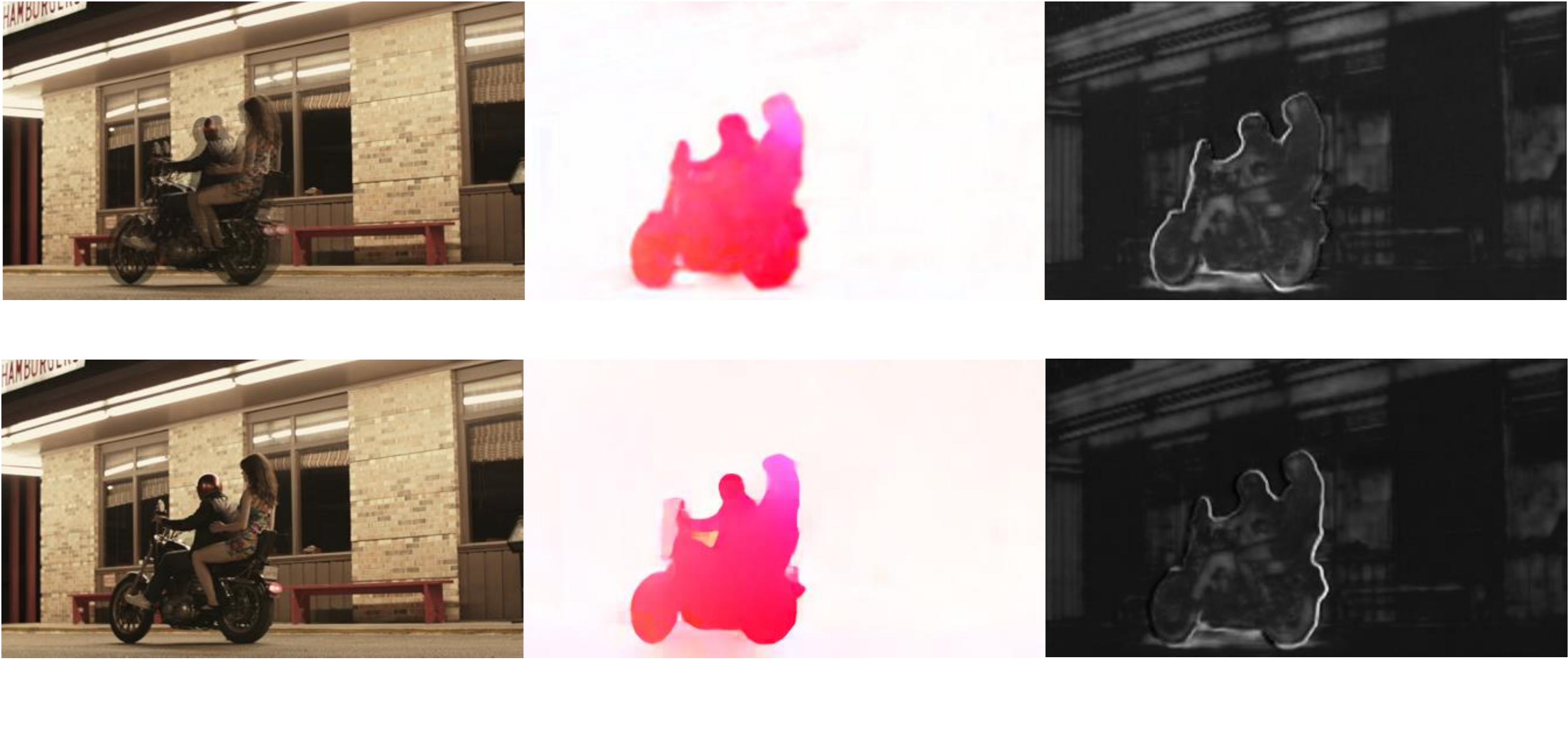}
        \put(10.5, 25){\footnotesize\linethickness{0.5mm}Overlaid}
        \put(39.5, 25){\footnotesize\linethickness{0.5mm}Our Avg. Flow}
        \put(72, 25){\footnotesize\linethickness{0.5mm}Variance ($t\rightarrow 0$)}
        \put(7, 2){\footnotesize\linethickness{0.5mm}Ground Truth}
        \put(38.5, 2){\footnotesize\linethickness{0.5mm}RAFT~\cite{teed2020raft} Flow}
        \put(72, 2){\footnotesize\linethickness{0.5mm}Variance ($t\rightarrow 1$)}
    \end{overpic}
    \caption{Visualizations of average and variance map of three flow pairs. 
    We provide RAFT~\cite{teed2020raft} flow for reference.
    }
    \label{fig:ab-flostat}
    \vspace{-15pt}
\end{figure}
\smallsec{Discussion.}
For further discussion, we visualize the mean and deviation of three estimated flow pairs. 
The results are shown in \figref{fig:ab-flostat}.
On the one hand, 
our average flow is generally consistent with the flow estimated from RAFT~\cite{teed2020raft},
which approximates to the ground truth displacements.
On the other hand, 
we observe that the major diversities of flows are at the motion boundaries and 
in the regions with rich textures.
This indicates that these regions need to involve more potential pixel candidates
 for reconstruction.
Through these visualizations,
we see that our method generate promising task-oriented flows,
\textit{generally consistent with the ground truth 
optical flows but diverse in local details}.

\vspace{-5pt}
\section{Conclusion}
Following the property of task-oriented flow, 
we have introduced All-pairs Multi-field Transforms (AMT) for efficient frame interpolation.
It contains two essential designs, including all-pairs correlation and multi-field refinement.
Through the two designs, our method could effectively handle large motions and occluded regions
during frame interpolation and achieve state-of-the-art performance on various benchmarks
with high efficiency.

\small{\noindent\textbf{Acknowledgment:}
This work is funded by the NSFC (NO. 62176130), Fundamental Research Funds for the Central Universities (Nankai University, NO. 63223050), China Postdoctoral Science Foundation (NO.2021M701780).
This work is partially supported by Ascend AI Computing Platform and CANN (Compute Architecture for Neural Networks).
We are also sponsored by CAAI-Huawei MindSpore Open Fund.}

{\small
\bibliographystyle{ieee_fullname}
\bibliography{egbib}
}

\appendix
\section*{Appendix}
\vspace{-3pt}
\section{Architecture Details}
\begin{figure*}[t]
    \centering
    \begin{overpic}[width=\textwidth]{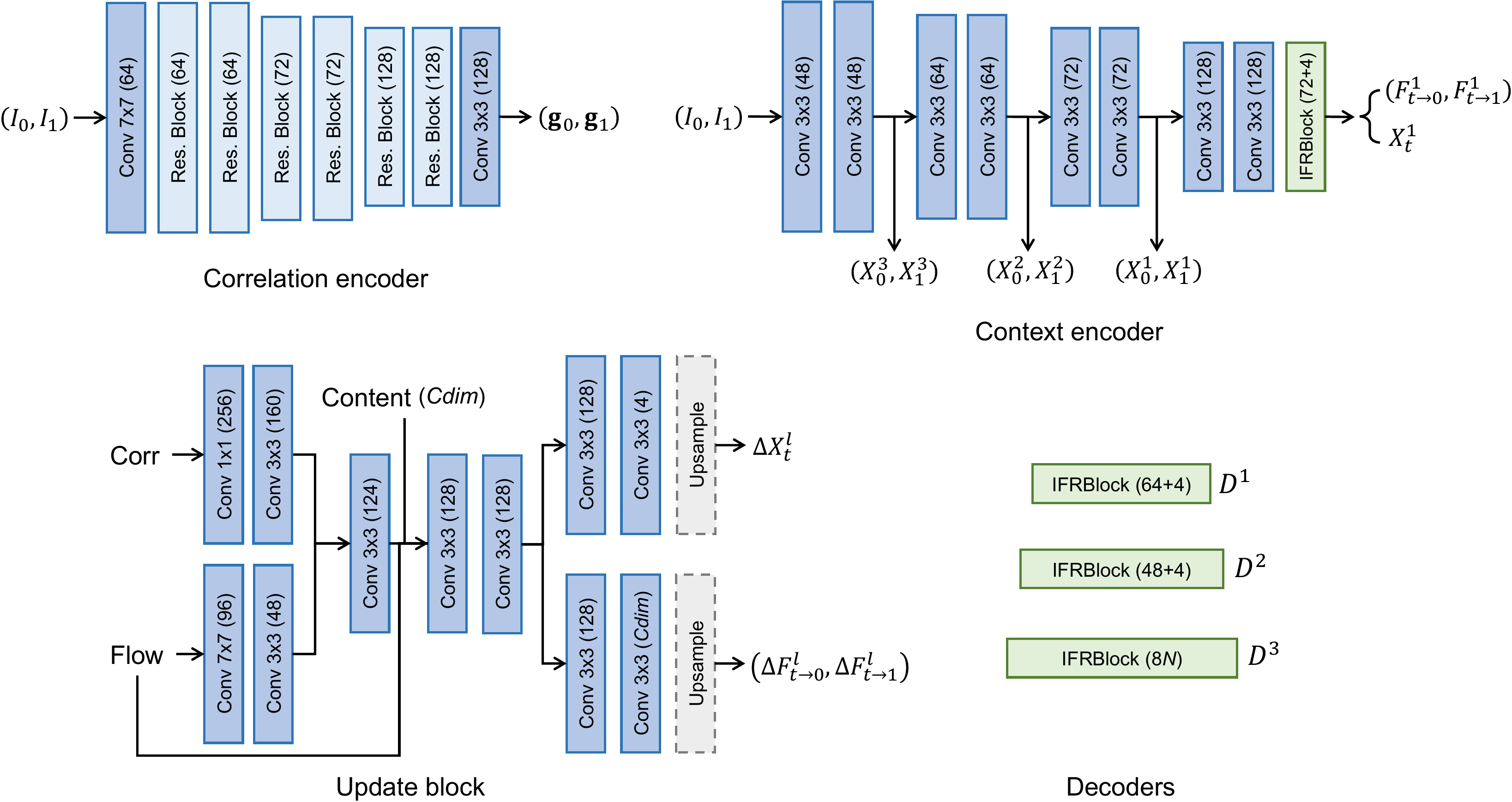}
    \end{overpic}
    \caption{Architecture details of the AMT-L. 
    The number in parentheses denotes the output channels.
    $N$ represents the number of output groups.
    IFRBlock denotes the decoder proposed in IFRNet~\cite{Kong_2022_CVPR}.
    }
    \label{fig:supp_amtl_arch}
\end{figure*}

\begin{figure*}[t]
    \centering
    \begin{overpic}[width=\textwidth]{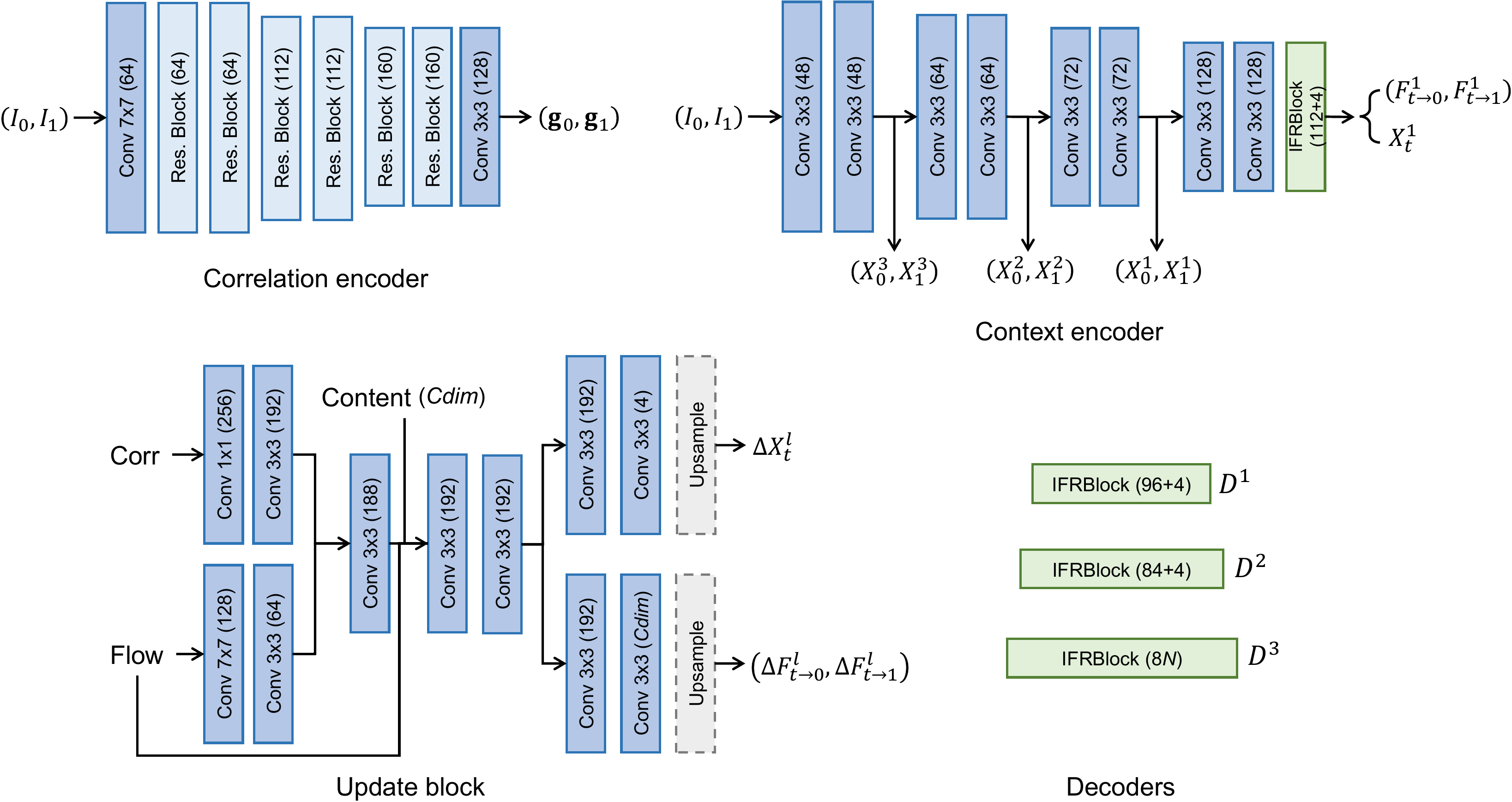}
    \end{overpic}
    \caption{Architecture details of the AMT-G. 
    The number in parentheses denotes the output channels.
    $N$ represents the number of output groups.
    IFRBlock denotes the decoder proposed in IFRNet~\cite{Kong_2022_CVPR}.
    }
    \label{fig:supp_amtg_arch}
\end{figure*}

\vspace{-3pt}
We build three models with different sizes, termed AMT-S, AMT-L, and AMT-G.
For reproducibility, the architecture details of them are shown in \figref{fig:supp_amts_arch}, \figref{fig:supp_amtl_arch}, and \figref{fig:supp_amtg_arch}, respectively.
We employ standard residual blocks~\cite{he2016deep} and instance normalization~\cite{ulyanov2016instance} in the correlation encoder.
The lookup radius is set to 3.
For each update block, a bilinear upsampling layer follows each head on upper levels (\ie $l \textgreater 1$).
The IFRBlock represents the decoder proposed in IFRNet~\cite{Kong_2022_CVPR},
which jointly estimates the bilateral flows and the intermediate feature.
To further improve performance, we upsample the correlation feature in the case of AMT-G to align its spatial resolution with the current interpolated feature, facilitating updates in the high-resolution space.
The code is available at \url{https://github.com/MCG-NKU/AMT}.

\vspace{-3pt}
\section{Multi-Frame Interpolation}
\vspace{-3pt}
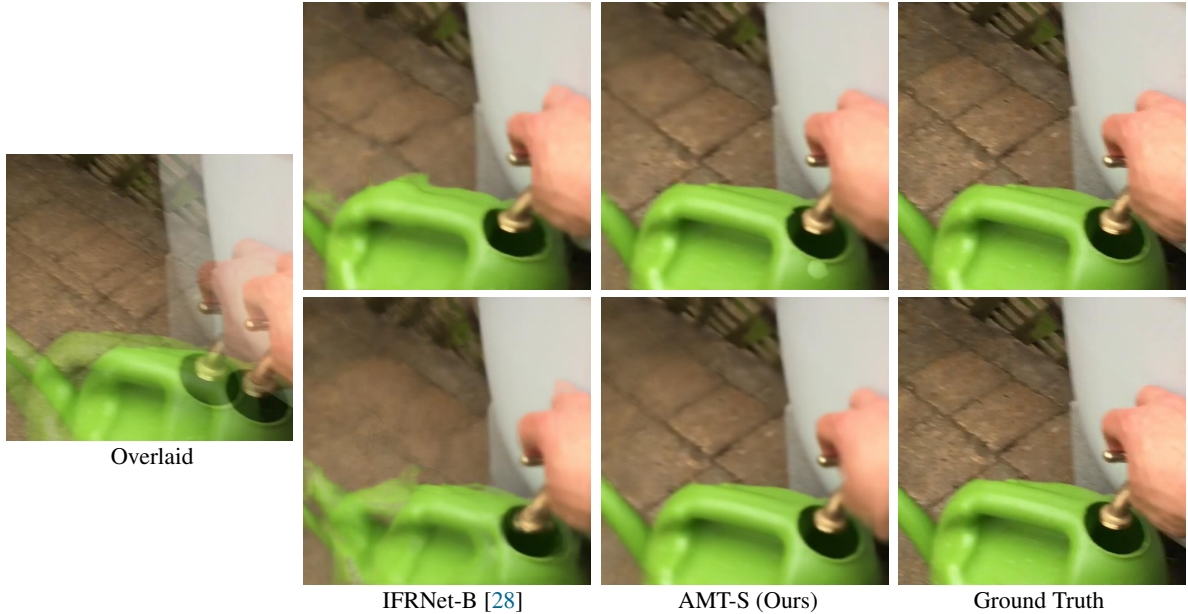
\begin{figure*}
    \centering
     \begin{overpic}[width=0.9\linewidth]{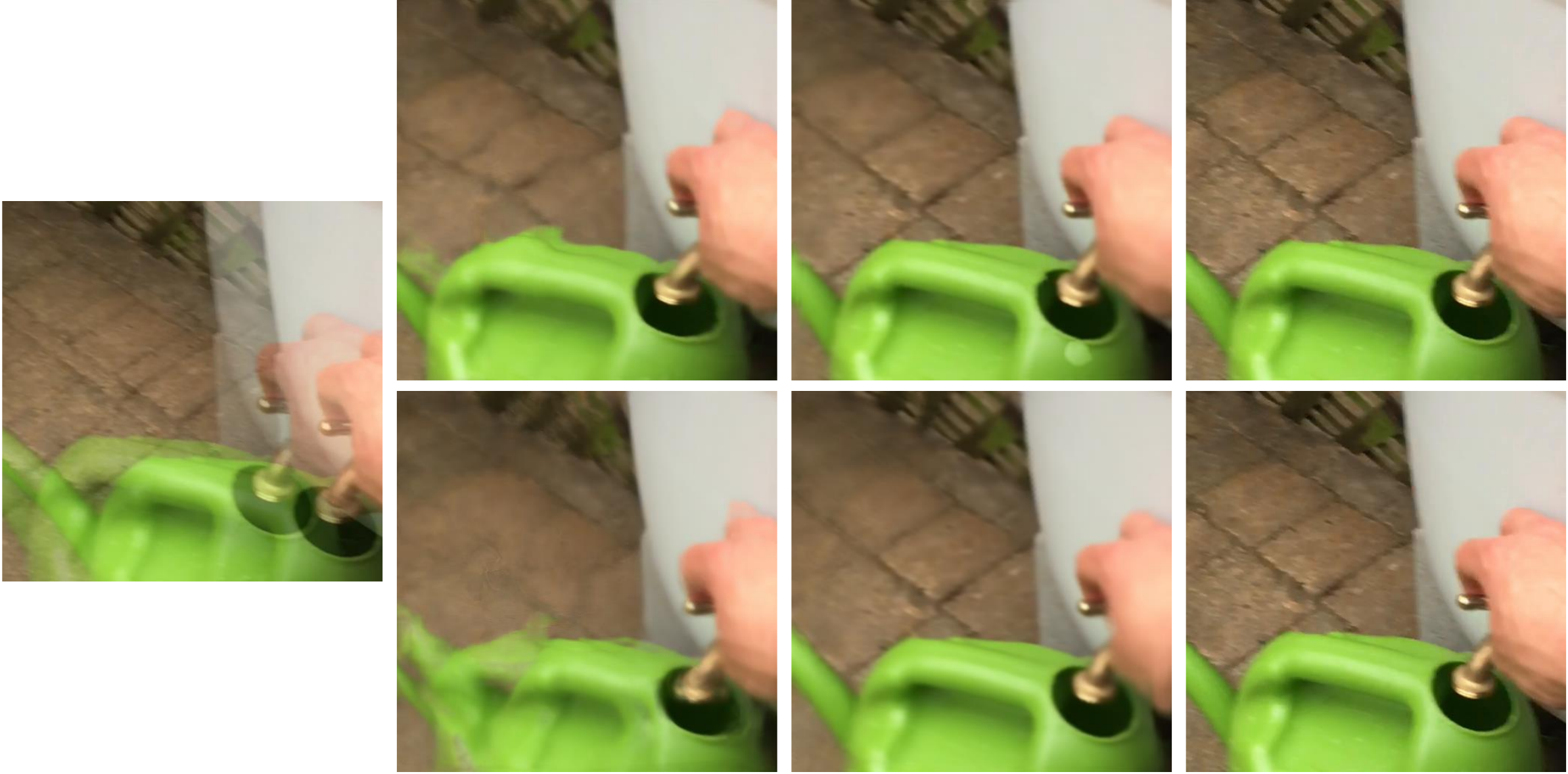}
        \put(9, 10.3){\small Overlaid}
        \put(32.0, -1.9){\small IFRNet-B~\cite{Kong_2022_CVPR}}
        \put(57, -1.9){\small AMT-S (Ours)}
        \put(82.0, -1.9){\small Ground Truth}
     \end{overpic}
     \vspace{1.5mm}
     \caption{Qualitative results of our AMT-S and IFRNet-B~\cite{Kong_2022_CVPR} on Adobe240~\cite{su2017deep}. The 
     time steps are $1/4$ and $1/2$ from top to bottom.}
     \label{fig:supp_comp_visual_adobe} 
     \vspace{-2mm}
 \end{figure*}

\begin{figure*}
    \centering
     \begin{overpic}[width=\linewidth]{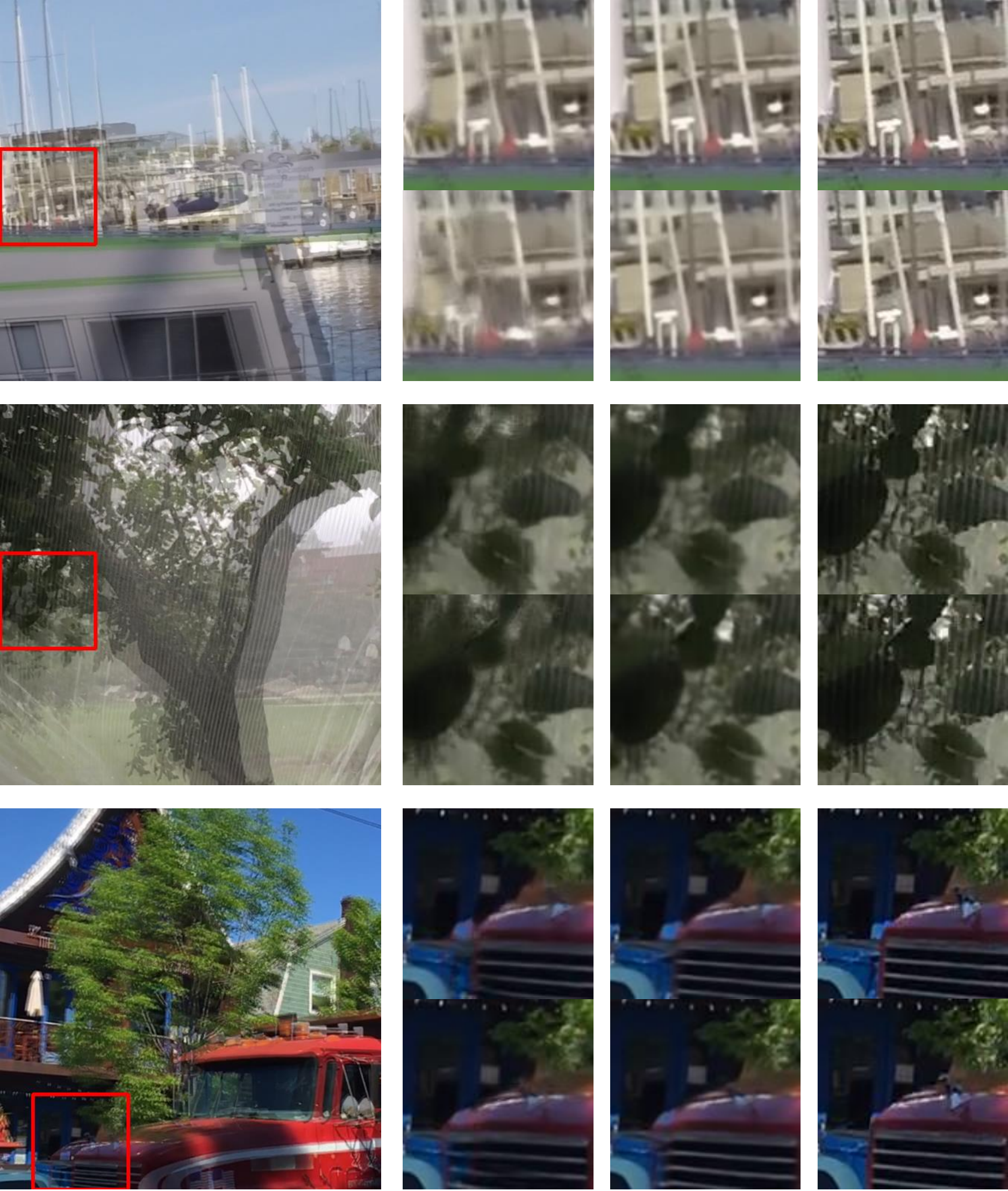}
        \put(14, 66.8){\small Overlaid}
        \put(38.0, 66.8){\small IFRNet-B~\cite{Kong_2022_CVPR}}
        \put(55., 66.8){\small AMT-S (Ours)}
        \put(72.5, 66.8){\small Ground Truth}
        \put(14, 32.8){\small Overlaid}
        \put(38.0, 32.8){\small IFRNet-B~\cite{Kong_2022_CVPR}}
        \put(55., 32.8){\small AMT-S (Ours)}
        \put(72.5, 32.8){\small Ground Truth}
        \put(14, -1.3){\small Overlaid}
        \put(38.0, -1.3){\small IFRNet-B~\cite{Kong_2022_CVPR}}
        \put(55., -1.3){\small AMT-S (Ours)}
        \put(72.5, -1.3){\small Ground Truth}
     \end{overpic}
     \vspace{1.5mm}
     \caption{Qualitative results of AMT-S and IFRNet-B~\cite{Kong_2022_CVPR} on Adobe240~\cite{su2017deep}. The 
     time steps are $1/4$ and $1/2$ from top to bottom.}
     \label{fig:supp_comp_visual_adobe2} 
 \end{figure*}

For the multi-frame setting, we use GoPro dataset~\cite{nah2017deep} for training 
and evaluate our model on the test partition of GoPro dataset~\cite{nah2017deep} and Adobe240 dataset~\cite{su2017deep}.
Here, we aim at $8 \times$ interpolation, 
synthesizing 7 intermediate frames with two input frames.
The other training settings and loss functions are consistent with those in our main paper.
Following recent frame interpolation works~\cite{huang2022rife, Kong_2022_CVPR},
we inject a temporal embedding vector into the network for $8 \times$ interpolation.
The elements in this vector are all set to $t$ according to the current time step, where $t \in \{ 1/8, 2/8, ..., 7/8\}$.
We compare our AMT-S with DVF~\cite{liu2017voxelflow}, SuperSloMo~\cite{jiang2018super}, DAIN~\cite{DAIN}, and IFRNet-B~\cite{Kong_2022_CVPR}.
The results of $8 \times$ interpolation are shown in \tabref{tab:comp_x8}.
Our method obtains the best PSNR and SSIM results on both evaluation datasets,
indicating the effectiveness of the proposed AMT for the task of multi-frame interpolation.
\figref{fig:supp_comp_visual_adobe} and \figref{fig:supp_comp_visual_adobe2} 
visually compare our method and IFRNet-B on the Adobe240 dataset.
Here, we visualize the cases for $1/4$ and $1/2$ time steps.
It can be seen that our method can generate more temporally consistent results
with fewer artifacts and more clear edges.

\begin{table}[t]
	\vspace{2mm}
	{\small
		\centering
        \tablestyle{6pt}{1.05}
		\begin{tabular}{ccccc}
			\toprule
			\multirow{2}[2]{*}{Method} & \multicolumn{2}{c}{GoPro~\cite{nah2017deep}} & \multicolumn{2}{c}{Adobe240~\cite{su2017deep}} \\
			\cmidrule(lr){2-3} \cmidrule(lr){4-5} & PSNR & SSIM & PSNR & SSIM \\
			\midrule
			DVF~\cite{liu2017voxelflow} & 21.94 & 0.776 & 28.23 & 0.896 \\
			SuperSloMo~\cite{jiang2018super} & 28.52 & 0.891 & 30.66 & 0.931 \\
			DAIN~\cite{DAIN} & 29.00 & 0.910 & 29.50 & 0.910 \\
			IFRNet-B~\cite{Kong_2022_CVPR} & 29.97 & 0.922 & 31.93 & 0.936 \\
			AMT-S (Ours) & \textbf{30.20} & \textbf{0.927} & \textbf{32.04} & \textbf{0.938}  \\
			\bottomrule
		\end{tabular}
		\vspace{-2mm}
		\caption{Quantitative comparison for 8$\times$ interpolation.}
		\label{tab:comp_x8}}
        \vspace{-2mm}
\end{table}

\section{Limitation}
Although our method has shown remarkable performance,
the 4D correlation volume computed from all pairs of pixels 
makes it hard to adapt to very high-resolution inputs
under a resource-constrained environment.
This is because
the computational complexity of constructing correlation volumes
is quadratic to the image resolution.
The possible ways to alleviate this problem include 
computing each correlation value only when it is looked up~\cite{teed2020raft}
or factorizing the 4D correlation volume to two 3D correlation volumes~\cite{xu2021high}.

\section{Discussions with RAFT}
Teed and Deng~\cite{teed2020raft} proposed RAFT,
which iteratively performs lookups on multi-scale 4D correlation volumes for updating flow fields.
Given its impressive results, 
current state-of-the-art flow estimation methods~\cite{jiang2021learning,xu2021high,Zhang2021SepFlow,xu2022gmflow,huang2022flowformer}
all derive from such architecture design.
Besides, it inspires the development of stereo matching~\cite{lipson2021raft} and scene flow~\cite{teed2021raft}.
However, the RAFT-like design paradigm is not well investigated in frame interpolation.

To better model large motions for frame interpolation, we build AMT based on RAFT. 
However, AMT involves many novel and task-specific designs beyond it.
To better illustrate our model, we detail the differences between our AMT and RAFT from the following perspectives:

\smallsec{Volume Design:}
RAFT constructs a unidirectional correlation volume
because it only needs to predict the optical flow along one direction.
For frame interpolation, we hope to model the dense correspondences on both directions
for updating bilateral flows.
We thus construct bidirectional correlation volumes.
We have verified the effectiveness of the bidirectional correlation volumes in Tab.~2a of the main paper. 

\smallsec{Context Encoder:}
In RAFT, the context encoder extracts the content feature only from the first input frame.
Because of the characteristics of frame interpolation,
in our AMT, the context encoder takes the image pair as input.
It outputs the initial intermediate feature,
the initial bilateral flows,
and the pyramid features from the input pair.
This design is also inspired by recent one-stage frame interpolation methods~\cite{Kong_2022_CVPR, reda2022film}.

\smallsec{Correlation Lookup:}
The lookup operation can be directly performed in RAFT 
for the identical coordinate system between the correlation volume and predicted flow field.
To solve the coordinate mismatch issue caused by the invisible frame,
we propose to scale the bilateral flows before the lookup operation.
Besides, 
we retrieve bidirectional correlations instead of the unidirectional ones in RAFT.
We use the initial bilateral flows $(F_{t\rightarrow 0}^{1}, F_{t\rightarrow 1}^{1})$
as the initial starting point, while RAFT uses zero instead.
The lookup strategy is investigated in Tab.2b of the main paper.

\smallsec{Predict and Update Manner:}
While RAFT predicts and updates the flow prediction at a single resolution,
we predict and update the bilateral flows in a coarse-to-fine manner.
We also provide a variant of our AMT to verify the design,
which only predicts the flow fields at a single resolution before feeding into the last decoder.
\tabref{tab:supp_raft_ablation} shows that this variant performs worse than the original one.
This indicates that predicting multi-scale flows are important for frame interpolation.
Besides, we also investigate the effectiveness of the cross-scale update in Tab.~2d of the main paper.

\begin{table}[t]
    \centering
    \tablestyle{4pt}{1.0}
    \begin{tabular}{lcccc}
        \toprule
        \multirow{2}[2]{*}{Case} & \multirow{2}[2]{*}{Vimeo90K~\cite{xue2019video}}  & \multicolumn{2}{c}{SNU-FILM~\cite{choi2020cain}} & \multirow{2}[2]{*}{FLOPs (G)} \\
        \cmidrule(lr){3-4} & & Hard & Extreme \\
        \midrule
        Single-scale Pred.   & 35.94 & 30.52 & 25.26 &  124 \\ 
        ConvGRU   & 35.99 & 30.58 & 25.27 & 132 \\ 
        Tied Weights   & 35.93 & 30.56 & 25.22 & 121 \\
        Convex Upsampling  & 35.99 & 30.56 & 25.28 & 123 \\
        Original Model    & \baseline{35.97} & \baseline{30.60} & \baseline{25.30} & 121 \\
        \bottomrule
    \end{tabular}
    \caption{Investigation on RAFT-like~\cite{teed2020raft} designs. 
    The default setting is marked in \colorbox{baselinecolor}{gray}.}
    \label{tab:supp_raft_ablation}
\end{table}

\smallsec{Update Block:}
In the design of the update block, our AMT differs from RAFT in five aspects:
1) While RAFT regards the feature extracted from the visible frame as the content guidance,
we use the interpolated intermediate feature representing the invisible frame instead.
2) RAFT only has one head in update block for regressing a flow residual,
while we have two heads for jointly predicting content and flow residuals.
The two aspects mentioned above have been discussed in Tab.~2c of the main paper. 
3) We stack two convolutional layers 
instead of a cumbersome ConvGRU unit in RAFT to handle the content and motion features.
We also investigate a variant that equips with a ConvGRU unit in each update block.
As shown in \tabref{tab:supp_raft_ablation},
this variant shows a comparable performance in contrast to the original one,
but it has more computational costs.
We thus choose to stack two convolutional layers for efficiency.
4) The weights of update blocks are not shared across levels in our AMT.
However, weight tying is beneficial to RAFT.
\tabref{tab:supp_raft_ablation} demonstrates that the model with untied weights performs better than that with tied weights.
5) We employ bilinear upsampling instead of convex sampling in RAFT for upscaling the flow fields.
As shown in \tabref{tab:supp_raft_ablation},
the two upsampling operators have similar performance,
but convex upsampling will incur more computation costs.
Thus, we choose the bilinear upsampling in our AMT.

\smallsec{Final Objective:}
RAFT is designed for flow estimation and is optimized only with flow regression loss.
However, our AMT is introduced for frame interpolation and is supervised 
with both task-oriented flow distillation loss and distortion-oriented content losses.
We need to  consider  not only the fidelity of estimated flows but also the diversity for meeting the requirement of task-oriented flows.
We thus output multiple flow pairs rather than a single flow field in RAFT.
Besides, occlusion reasoning and residual hallucination also need to be considered
for faithful content generation.

\section{Discussion about Multi-Field Refinement}
Some works~\cite{BMBC, park2021ABME,hu2022m2m} also attempt to predict multiple flow pairs
for preparing intermediate content candidates.
Specifically, BMBC~\cite{BMBC} predicts six bilateral motions 
through the bilateral motion network and optical flow approximation.
ABME~\cite{park2021ABME} generates four bilateral flow fields based on asymmetric motion assumption.
After obtaining warped candidate frames and context features,
the two works rely on a dynamic filter and even a cumbersome synthesis network
to generate the final intermediate frame.
Thus, they are inefficient for practical usage.
In contrast, our AMT is more efficient, as shown in Tab.~1 of the main paper.
We generate multiple flow fields in a single forward pass instead of multiple inference steps in BMBC and ABME.
Besides, we obtain the intermediate candidates only in the image domain rather than the feature domain
 and stack two lightweight convolutional layers for fusing these candidates.

M2M-VFI~\cite{hu2022m2m} is most relevant to our multi-field refinement.
It also generates multiple flows in one step and prepares warped candidates in the image domain.
However, there are five key differences between our multi-field refinement and M2M-VFI.  
First, our method generates the candidate frames by backward warping rather than forward warping in M2M-VFI.
Second, while M2M-VFI predicts multiple flows to 
overcome the hole issue and artifacts in overlapped regions caused by forward warping,
we aim to alleviate the ambiguity issue in the occluded areas and motion boundaries 
 by enhancing the diversity of flows.
Third, M2M-VFI needs to estimate bidirectional flows first through an off-the-shelf optical flow estimator
and then predict multiple bilateral flows through a motion refinement network.
On the contrary, we directly estimate multiple bilateral flows in a one-stage network.
In this network, we first estimate one pair of bilateral flows at the coarse scale 
and then derive multiple groups of fine-grained bilateral flows from the coarse flow pairs.
Fourth, 
M2M-VFI jointly estimates two reliability maps together with all pairs of bilateral flows,
which can be further used to fuse the overlapping pixels caused by forward warping.
As shown in Eqn.~(5) of the main paper, 
we estimate not only an occlusion mask but a residual content for cooperating with each pair of bilateral flows.
The residual content is used to compensate for the unreliable details after warping.
This design has been investigated in Tab.~2e of the main paper.
Fifth, we stack two convolutional layers to adaptively merge candidate frames, 
while M2M-VFI normalizes the sum of all candidate frames through a pre-computed weighting map.

\section{More Visual Results}
\begin{figure}[t]
    \begin{overpic}[width=\linewidth]{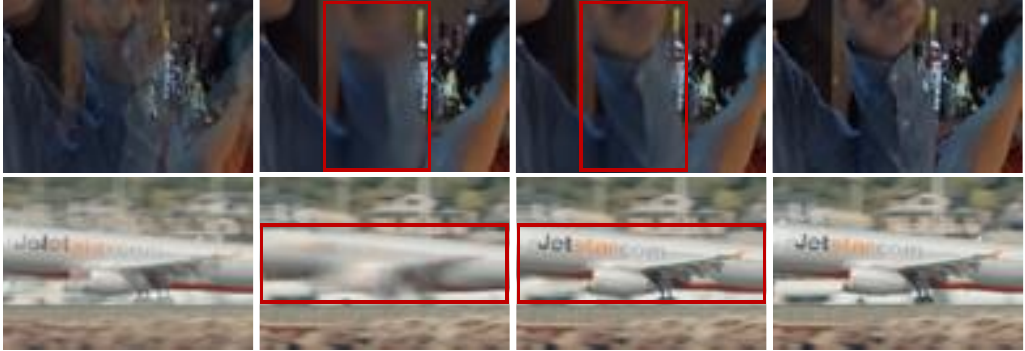}
        \put(5.4, -3.00){\small\linethickness{0.5mm}Overlaid}
        \put(27.5, -3.00){\small\linethickness{0.5mm}VFIFormer}
        \put(51.2, -3.00){\small\linethickness{0.5mm}AMT-G (Ours)}
        \put(77.5, -3.00){\small\linethickness{0.5mm}Ground Truth}
    \end{overpic}
    \caption{Qualitative comparison between AMT-G with VFIFormer. Our method recovers more clear structure and edges.
    }
    \label{fig:amt-g}
\end{figure}

\begin{figure*}
    \centering
    \begin{overpic}[width=0.96\linewidth]{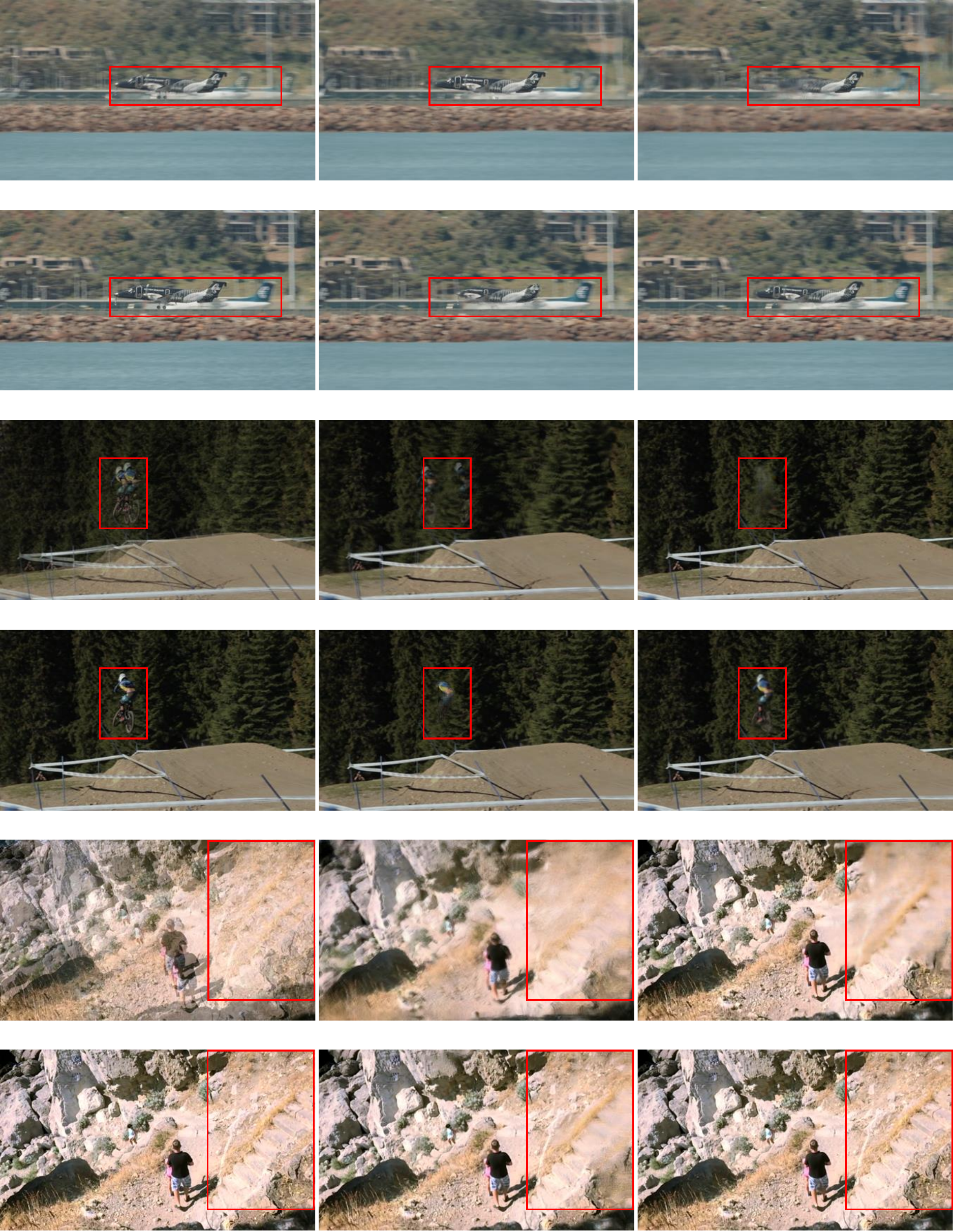}
       \put(10, 84.){\small Overlaid}
       \put(35, 84.){\small AdaCoF~\cite{lee2020adacof}}
       \put(62, 84.){\small RIFE~\cite{huang2022rife}}
       \put(9, -1.3){\small Ground Truth}
       \put(9, 32.8){\small Ground Truth}
       \put(9, 67){\small Ground Truth}
       \put(10, 15.8){\small Overlaid}
       \put(10, 50){\small Overlaid}
       \put(35, -1.3){\small IFRNet-B~\cite{Kong_2022_CVPR}}
       \put(35, 32.8){\small IFRNet-B~\cite{Kong_2022_CVPR}}
       \put(35, 67){\small IFRNet-B~\cite{Kong_2022_CVPR}}
       \put(60.5, -1.3){\small AMT-S (Ours)}
       \put(60.5, 32.8){\small AMT-S (Ours)}
       \put(60.5, 67){\small AMT-S (Ours)}
       \put(35, 15.8){\small AdaCoF~\cite{lee2020adacof}}
       \put(35, 50.0){\small AdaCoF~\cite{lee2020adacof}}
       \put(62, 15.8){\small RIFE~\cite{huang2022rife}}
       \put(62, 50.0){\small RIFE~\cite{huang2022rife}}
    \end{overpic}
    \vspace{1.5mm}
    \caption{Visual comparison for the methods with low computational complexity on Vimeo90K dataset~\cite{xue2019video}}
    \label{fig:supp_comp_visual_1} 
\end{figure*}

\begin{figure*}
    \centering
    \begin{overpic}[width=0.97\linewidth]{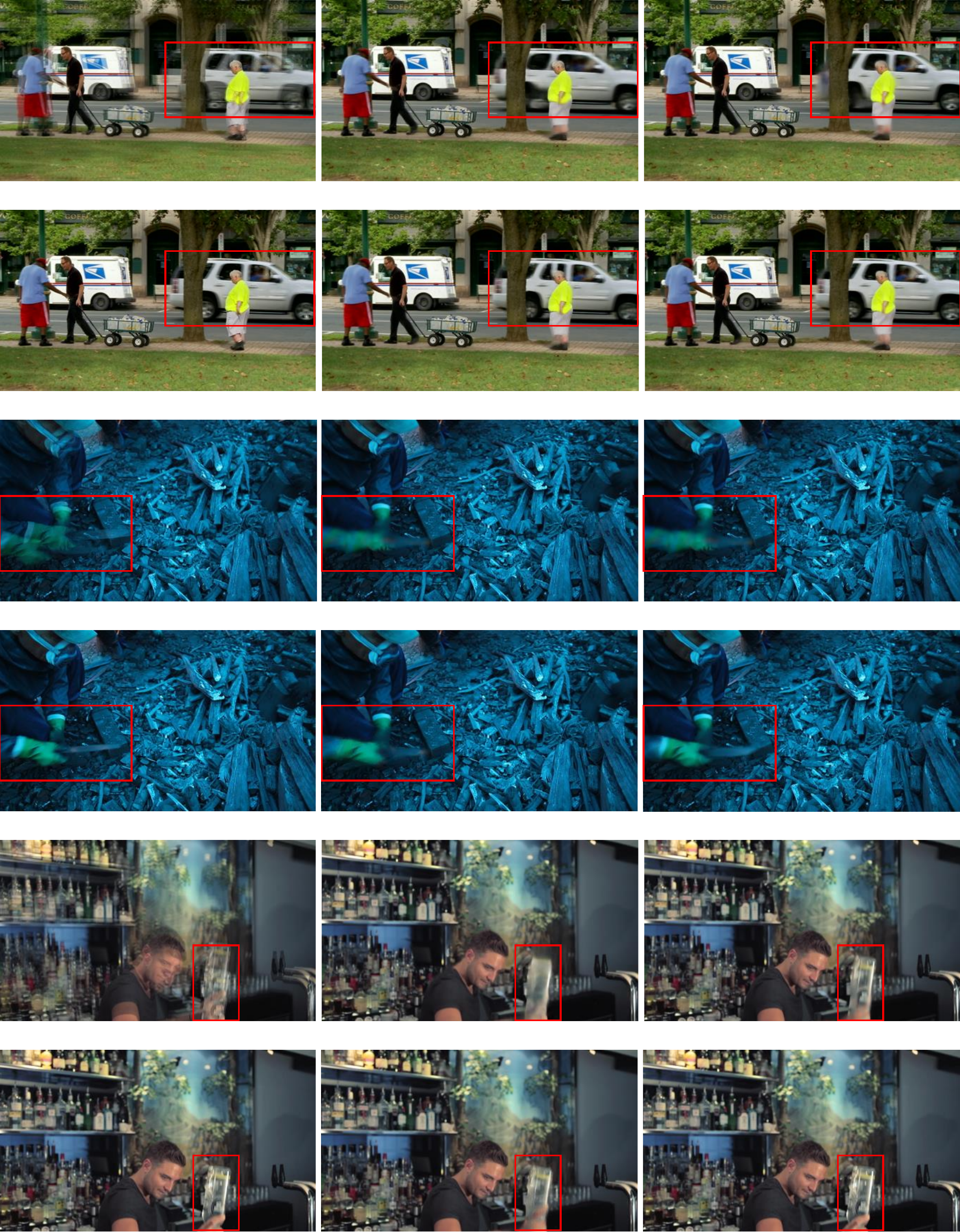}
       \put(9.2, -1.5){\small Ground Truth}
       \put(9.2, 32.8){\small Ground Truth}
       \put(9.2, 66.8){\small Ground Truth}
       \put(10.2, 15.6){\small Overlaid}
       \put(10.2, 49.7){\small Overlaid}
       \put(10.2, 83.8){\small Overlaid}
       \put(35.1, -1.5){\small IFRNet-L~\cite{Kong_2022_CVPR}}
       \put(35.1, 32.8){\small IFRNet-L~\cite{Kong_2022_CVPR}}
       \put(35.1, 66.8){\small IFRNet-L~\cite{Kong_2022_CVPR}}
       \put(60.8, -1.5){\small AMT-L (Ours)}
       \put(60.8, 32.8){\small AMT-L (Ours)}
       \put(60.8, 66.8){\small AMT-L (Ours)}
       \put(36.2, 15.6){\small CAIN~\cite{choi2020cain}}
       \put(36.2, 49.7){\small CAIN~\cite{choi2020cain}}
       \put(36.2, 83.8){\small CAIN~\cite{choi2020cain}}
       \put(61.5, 15.6){\small ABME~\cite{park2021ABME}}
       \put(61.5, 49.7){\small ABME~\cite{park2021ABME}}
       \put(61.5, 83.8){\small ABME~\cite{park2021ABME}}
    \end{overpic}
    \vspace{1.5mm}
    \caption{Visual comparison for the methods with relatively high computational complexity on Vimeo90K dataset~\cite{xue2019video}.}
    \label{fig:supp_comp_visual_2} 
\end{figure*}

\begin{figure*}
    \centering
    \begin{overpic}[width=\linewidth]{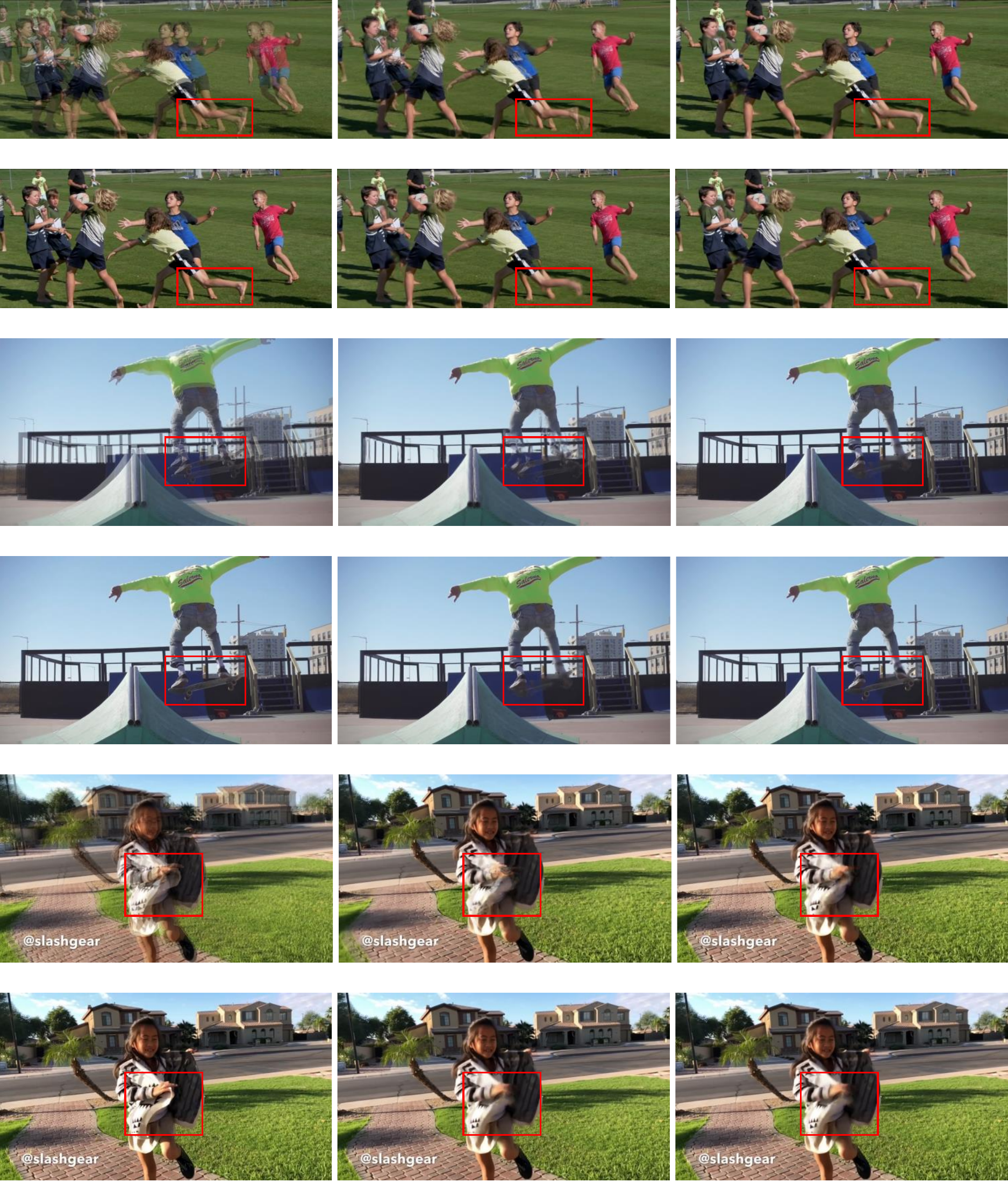}
      \put(9, -1.3){\small Ground Truth}
      \put(9, 35.6){\small Ground Truth}
      \put(9, 72.4){\small Ground Truth}
      \put(10.5, 17.2){\small Overlaid}
      \put(10.5, 54){\small Overlaid}
      \put(10.5, 86.8){\small Overlaid}
      \put(38.45, -1.3){\small IFRNet-B~\cite{Kong_2022_CVPR}}
      \put(38.45, 35.6){\small IFRNet-B~\cite{Kong_2022_CVPR}}
      \put(38.45, 72.4){\small IFRNet-B~\cite{Kong_2022_CVPR}}
      \put(66.7, -1.3){\small AMT-S (Ours)}
      \put(66.7, 35.6){\small AMT-S (Ours)}
      \put(66.7, 72.4){\small AMT-S (Ours)}
      \put(38.7, 17.2){\small AdaCoF~\cite{lee2020adacof}}
      \put(38.7, 54){\small AdaCoF~\cite{lee2020adacof}}
      \put(38.7, 86.8){\small AdaCoF~\cite{lee2020adacof}}
      \put(68.7, 17.2){\small RIFE~\cite{huang2022rife}}
      \put(68.7, 54){\small RIFE~\cite{huang2022rife}}
      \put(68.7, 86.8){\small RIFE~\cite{huang2022rife}}
    \end{overpic}
    \vspace{1.5mm}
    \caption{Visual comparison for the methods with low computational complexity on the Hard partition in SNU-FILM dataset~\cite{choi2020cain}.}
    \label{fig:supp_comp_visual_3} 
\end{figure*}

\begin{figure*}
    \centering
    \begin{overpic}[width=\linewidth]{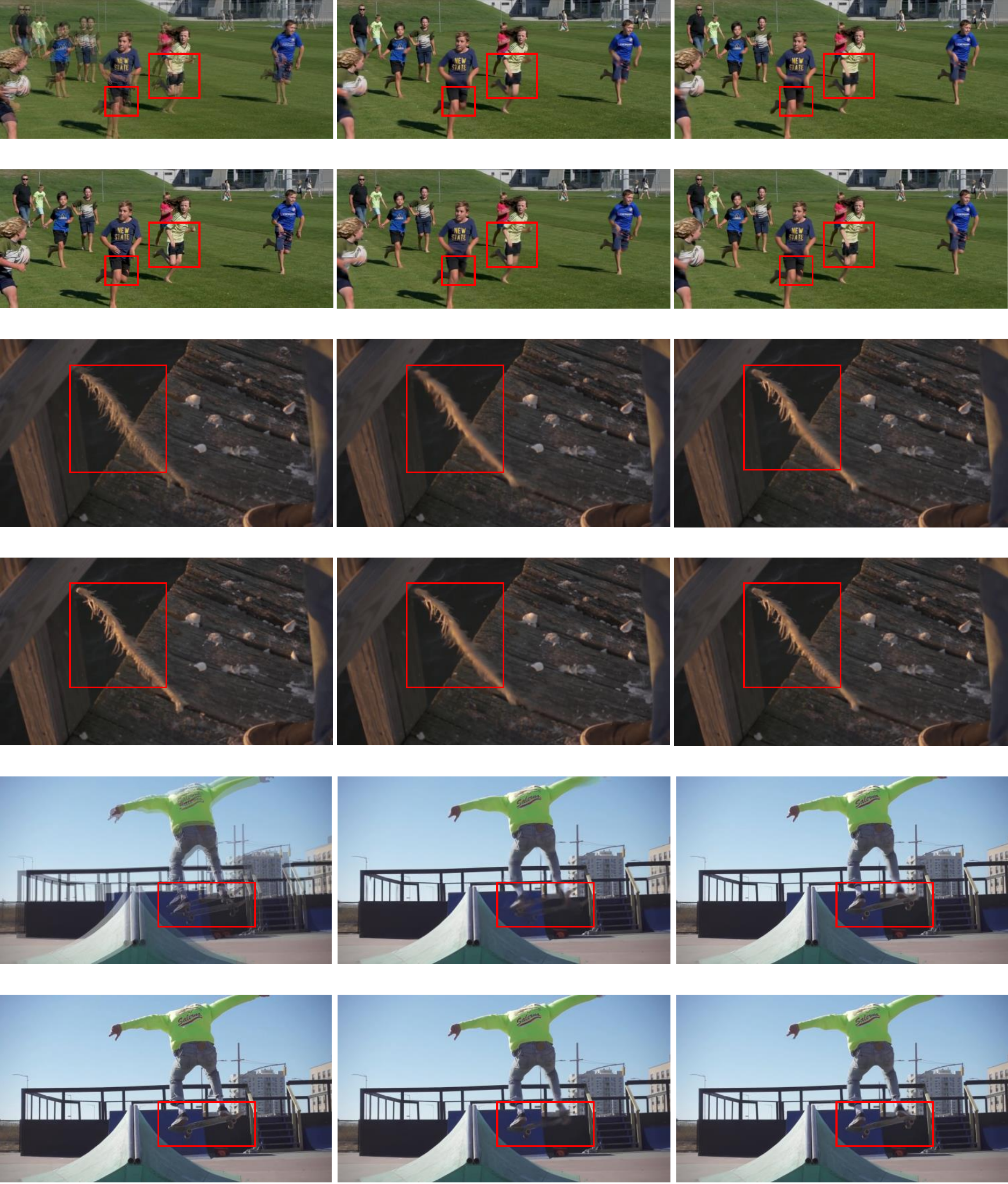}
      \put(9.3, -1.3){\small Ground Truth}
      \put(9.3, 35.6){\small Ground Truth}
      \put(9.3, 72.4){\small Ground Truth}
      \put(10.5, 17.1){\small Overlaid}
      \put(10.5, 54){\small Overlaid}
      \put(10.5, 86.7){\small Overlaid}
      \put(38.6, -1.3){\small IFRNet-L~\cite{Kong_2022_CVPR}}
      \put(38.6, 35.6){\small IFRNet-L~\cite{Kong_2022_CVPR}}
      \put(38.6, 72.4){\small IFRNet-L~\cite{Kong_2022_CVPR}}
      \put(66.7, -1.3){\small AMT-L (Ours)}
      \put(66.7, 35.6){\small AMT-L (Ours)}
      \put(66.7, 72.4){\small AMT-L (Ours)}
      \put(39.6, 17.1){\small CAIN~\cite{choi2020cain}}
      \put(39.6, 54){\small CAIN~\cite{choi2020cain}}
      \put(39.6, 86.7){\small CAIN~\cite{choi2020cain}}
      \put(67.5, 17.1){\small ABME~\cite{park2021ABME}}
      \put(67.5, 54){\small ABME~\cite{park2021ABME}}
      \put(67.5, 86.7){\small ABME~\cite{park2021ABME}}
    \end{overpic}
    \vspace{1.5mm}
    \caption{Visual comparison for the methods with relatively high computational complexity on the Hard partition in SNU-FILM dataset~\cite{choi2020cain}.}
    \label{fig:supp_comp_visual_5} 
\end{figure*}

\begin{figure*}
    \centering
    \begin{overpic}[width=\linewidth]{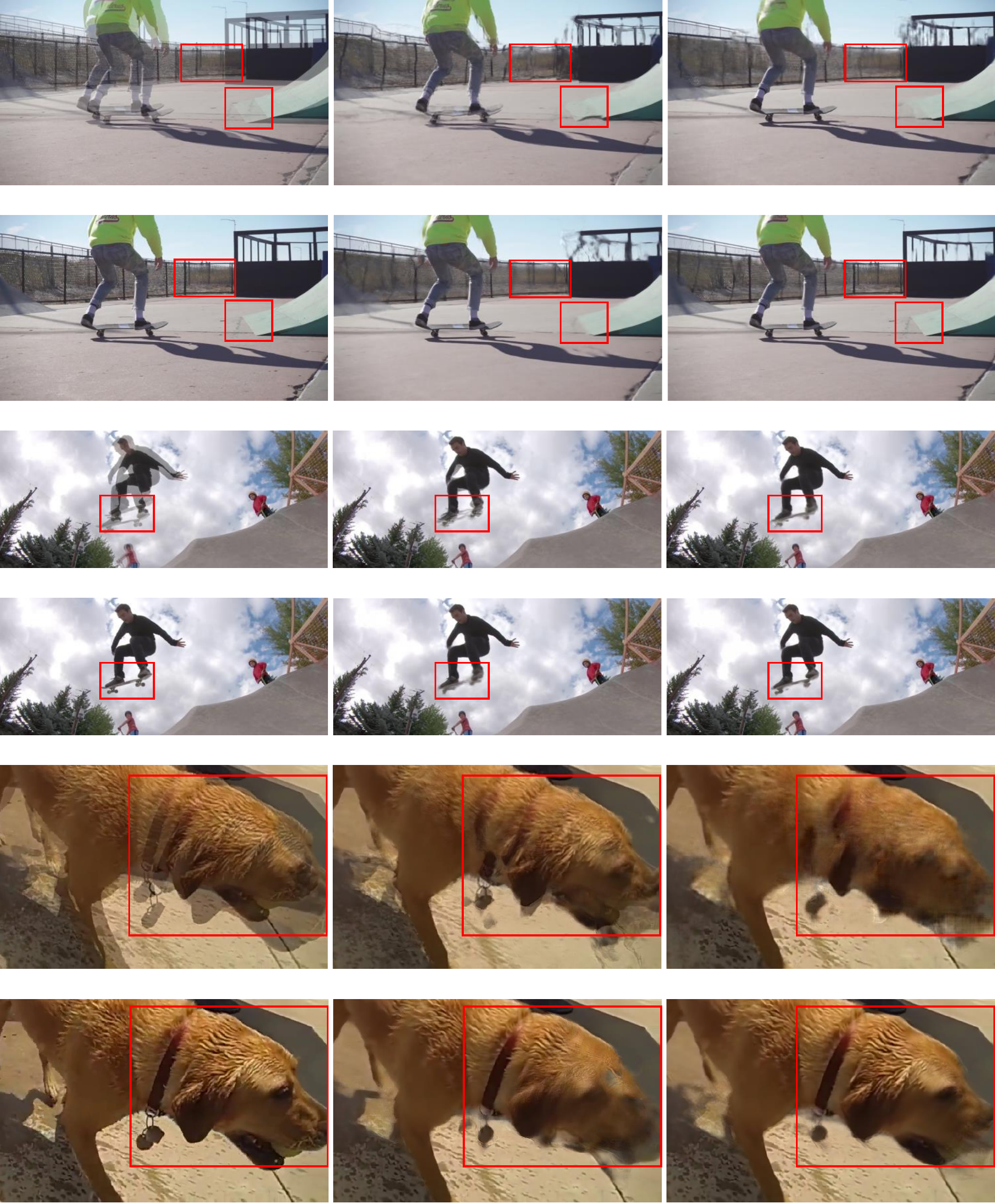}
      \put(8.5, -1.3){\small Ground Truth}
      \put(8.5, 37.6){\small Ground Truth}
      \put(8.5, 65.3){\small Ground Truth}
      \put(10., 18.2){\small Overlaid}
      \put(10., 51.5){\small Overlaid}
      \put(10., 83.1){\small Overlaid}
      \put(37.4, -1.3){\small IFRNet-B~\cite{Kong_2022_CVPR}}
      \put(37.4, 37.6){\small IFRNet-B~\cite{Kong_2022_CVPR}}
      \put(37.4, 65.3){\small IFRNet-B~\cite{Kong_2022_CVPR}}
      \put(64.7, -1.3){\small AMT-S (Ours)}
      \put(64.7, 37.6){\small AMT-S (Ours)}
      \put(64.7, 65.3){\small AMT-S (Ours)}
      \put(37.6, 18.2){\small AdaCoF~\cite{lee2020adacof}}
      \put(37.6, 51.5){\small AdaCoF~\cite{lee2020adacof}}
      \put(37.6, 83.1){\small AdaCoF~\cite{lee2020adacof}}
      \put(66.7, 18.2){\small RIFE~\cite{huang2022rife}}
      \put(66.7, 51.5){\small RIFE~\cite{huang2022rife}}
      \put(66.7, 83.1){\small RIFE~\cite{huang2022rife}}
    \end{overpic}
    \vspace{1.5mm}
    \caption{Visual comparison for the methods with low computational complexity on the Extreme partition in SNU-FILM dataset~\cite{choi2020cain}.}
    \label{fig:supp_comp_visual_4} 
\end{figure*}

\begin{figure*}
    \centering
    \begin{overpic}[width=\linewidth]{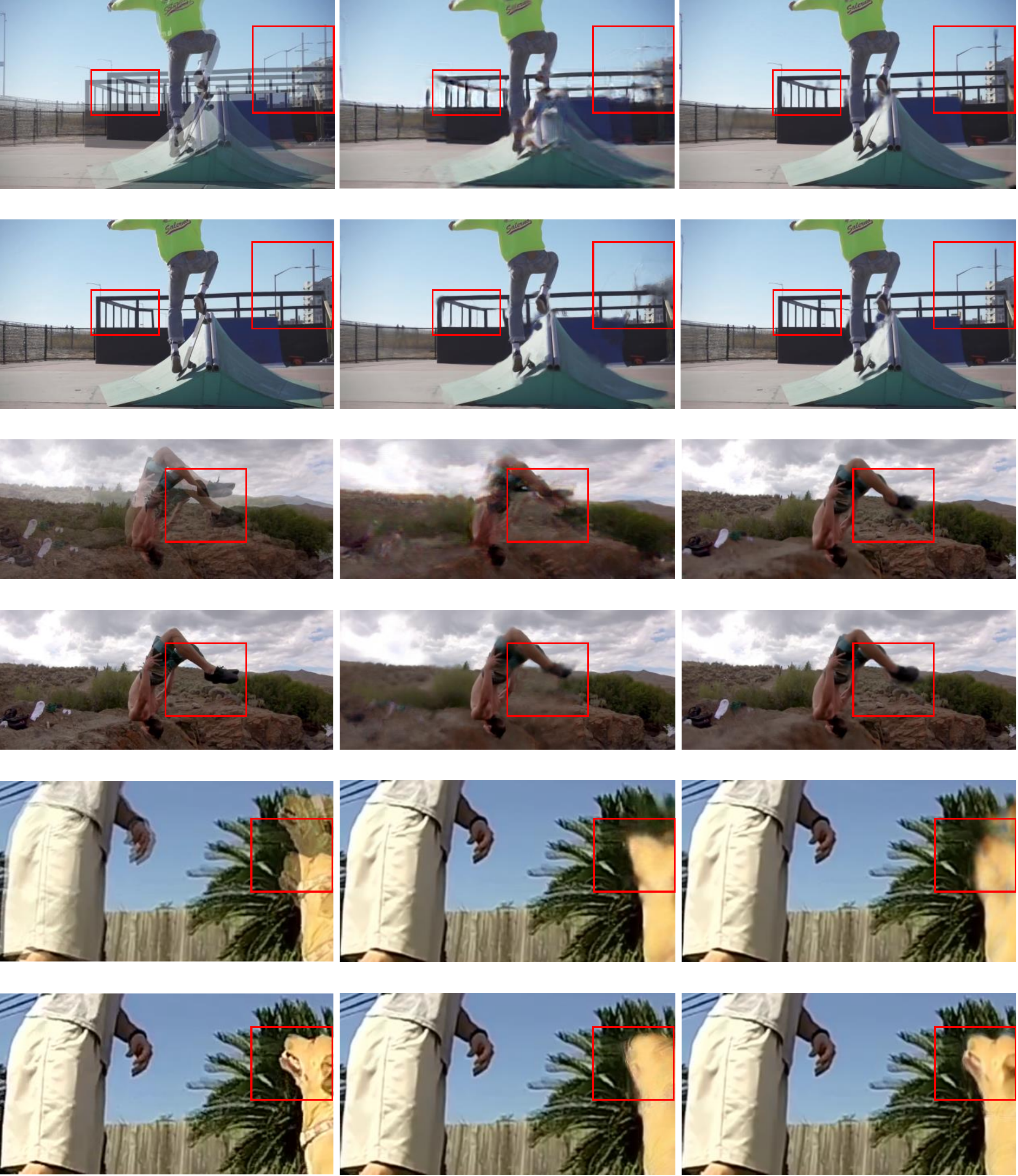}
      \put(9.8, -1.5){\small Ground Truth}
      \put(9.8, 34.8){\small Ground Truth}
      \put(9.8, 63.8){\small Ground Truth}
      \put(10.9, 16.7){\small Overlaid}
      \put(10.9, 49.3){\small Overlaid}
      \put(10.9, 82.4){\small Overlaid}
      \put(38.9, -1.5){\small IFRNet-L~\cite{Kong_2022_CVPR}}
      \put(38.9, 34.8){\small IFRNet-L~\cite{Kong_2022_CVPR}}
      \put(38.9, 63.8){\small IFRNet-L~\cite{Kong_2022_CVPR}}
      \put(67.4, -1.5){\small AMT-L (Ours)}
      \put(67.4, 34.8){\small AMT-L (Ours)}
      \put(67.4, 63.8){\small AMT-L (Ours)}
      \put(40.2, 16.7){\small CAIN~\cite{choi2020cain}}
      \put(40.2, 49.3){\small CAIN~\cite{choi2020cain}}
      \put(40.2, 82.4){\small CAIN~\cite{choi2020cain}}
      \put(68.2, 16.7){\small ABME~\cite{park2021ABME}}
      \put(68.2, 49.3){\small ABME~\cite{park2021ABME}}
      \put(68.2, 82.4){\small ABME~\cite{park2021ABME}}
    \end{overpic}
    \vspace{1.5mm}
    \caption{Visual comparison for the methods with relatively high computational complexity on the Extreme partition in SNU-FILM dataset~\cite{choi2020cain}.}
    \label{fig:supp_comp_visual_6} 
\end{figure*}

In this section, 
we provide additional visual results on two benchmark datasets, 
including Vimeo90K~\cite{xue2019video} and SNU-FILM~\cite{choi2020cain},
to further show the superiority of the proposed AMT.
The comparison methods include CAIN~\cite{choi2020cain}, AdaCoF~\cite{lee2020adacof}, 
ABME~\cite{park2021ABME}, RIFE~\cite{huang2022rife}, 
IFRNet(-B/-L)~\cite{Kong_2022_CVPR}, and VFIFormer~\cite{lu2022vfiformer}.
For a fair comparison, we also divide these methods into two groups
according to the computational cost.
As shown in \figref{fig:amt-g}, \ref{fig:supp_comp_visual_1}-\ref{fig:supp_comp_visual_6},
our AMT synthesizes the object with large motions more faithfully and 
generates plausible textures with fewer artifacts. 

\section{Broader Impact}
As presented in this paper,
our AMT can synthesize faithful non-existent frames between two visible frames.
Given its reliable synthesis results, our method may be abused to forge or tamper with videos.

\end{document}